%% file: main.tex
\documentclass[10pt,twocolumn,letterpaper]{article}

\usepackage{cvpr}              
\usepackage{algorithm}
\usepackage{algorithmic}
\usepackage{array}
\usepackage{multirow}
\usepackage{graphicx}
\usepackage{xcolor}
\usepackage{xspace}
\usepackage{amssymb}
\usepackage{amsmath} 
\usepackage{amsmath, amsthm, amssymb}

\newtheorem{assumption}{Assumption}

\newtheorem{theorem}{Theorem}

\newcommand{\coloneqq}{\mathrel{:=}}
\DeclareMathOperator*{\argmax}{arg\,max}

\usepackage{booktabs} 
\input{preamble}
\definecolor{cvprblue}{rgb}{0.21,0.49,0.74}
\usepackage[pagebackref,breaklinks,colorlinks,allcolors=cvprblue]{hyperref}

\def\paperID{} 
\def\confName{CVPR}
\def\confYear{2026}

\definecolor{mygray}{gray}{.92}
\definecolor{ForestGreen}{RGB}{34,139,34}
\definecolor{Forestred}{RGB}{220,50,50}

\newcommand{\algname}{Prune2Drive\xspace} 

\title{Prune2Drive: A Plug-and-Play Framework for Accelerating Vision-Language Models in Autonomous Driving}

\author{
    Minhao Xiong\textsuperscript{\rm 1}\thanks{Equal Contribution.} \quad
    Zichen Wen\textsuperscript{\rm 1,2}\footnotemark[1]  \quad
    Zhuangcheng Gu\textsuperscript{\rm 3}  \quad
    Xuyang Liu\textsuperscript{\rm 4}  \\
    Rui Zhang\textsuperscript{\rm 2}  \quad
    Hengrui Kang\textsuperscript{\rm 1,2} \quad
    Jiabing Yang\textsuperscript{\rm 5} \quad
    Junyuan Zhang\textsuperscript{\rm 6,2}  \\
    Weijia Li\textsuperscript{\rm 2}  \quad
    Conghui He\textsuperscript{\rm 2} \quad
    Yafei Wang\textsuperscript{\rm 1} \quad
    Linfeng Zhang\textsuperscript{\rm 1}\thanks{Corresponding author (zhanglinfeng@sjtu.edu.cn).}\\
    \textsuperscript{\rm 1}Shanghai Jiao Tong University \quad 
    \textsuperscript{\rm 2}Shanghai AI Laboratory \quad
    \textsuperscript{\rm 3}Carnegie Mellon University \\
    \textsuperscript{\rm 4}Sichuan University \quad
    \textsuperscript{\rm 5}University of Chinese Academy of Sciences \quad
    \textsuperscript{\rm 6}The University of Hong Kong 
}

\begin{document}
\maketitle
{
    \hypersetup{linkcolor=black}
}
\input{contents/0_abstract}

\input{contents/1_introduction_wzc}

\input{contents/2_related_works}

\input{contents/3_method}

\input{contents/4_experiments}

\input{contents/5_conclusion}

{
    \small
    \bibliographystyle{ieeenat_fullname}
    \bibliography{main}
}

\input{contents/x_supp}

\end{document}

%% file: contents/0_abstract.tex
\begin{abstract}
Vision-Language Models (VLMs) have emerged as a promising paradigm in autonomous driving (AD), providing a unified framework for perception and decision-making. 
However, their real-world deployment is hindered by significant computational overhead when processing high-resolution, multi-view images. 
This complexity stems from the massive number of visual tokens, which increases inference latency and memory consumption due to the quadratic complexity of self-attention.
To address these challenges, we propose \textbf{\algname}, a plug-and-play visual token pruning framework for multi-view VLMs in AD. 
\algname introduces two core innovations: (i) a diversity-aware token selection mechanism that prioritizes semantic and spatial coverage across views, and (ii) a view-adaptive pruning controller that automatically learns optimal pruning ratios based on camera importance to downstream tasks. 
Unlike prior methods, \algname requires no model retraining or access to attention maps, ensuring compatibility with modern efficient attention implementations.
Extensive experiments on the DriveLM and DriveLMM-o1 benchmarks demonstrate that \algname achieves significant speedups and memory savings with minimal performance impact. 
When retaining only 10\% of visual tokens, our method achieves a 6.40$\times$ speedup in the prefilling phase and consumes only 13.4\% of the original FLOPs, with a mere 3\% average performance drop on the DriveLM benchmark. 
Code is available at: \href{https://github.com/MinhaoXiong/Prune2Drive.git}{https://github.com/MinhaoXiong/Prune2Drive.git}

\end{abstract}

%% file: contents/1_introduction_wzc.tex
\section{Introduction}\label{sec:introduction}
\begin{figure}[!t]
    \centering
    \includegraphics[width=\linewidth]{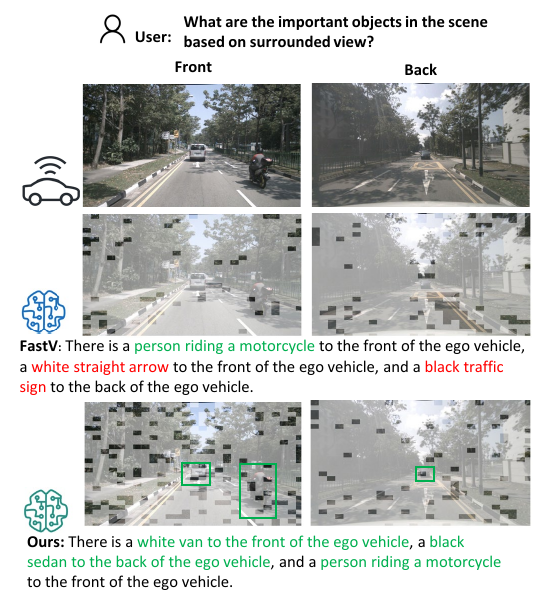}
    \vspace{-5mm}
    \caption{Comparison of the front-view and back-view visual token selection between \algname and FastV. The green box highlights important objects captured by \algname, the red text wrongly describes the scene, and the green text matches the scene. 
    }
    \label{fig:teaser} 

    \vspace*{-0.6cm}
\end{figure}
Vision-Language Models (VLMs)~\cite{li2024llava,chen2024internvl,team2025kimi, team2026kimi, wen2026innovator} have emerged as a promising paradigm in autonomous driving (AD), seamlessly integrating visual perception with linguistic reasoning. By enabling natural language interactions grounded in visual contexts, VLMs offer enhanced interpretability, more intuitive decision-making, and a unified framework for solving complex driving tasks, such as detection, prediction, and planning~\cite{zhou2024vision, xu2024vlm, jiang2024sennabridginglargevisionlanguage, kim2018textualexplanationsselfdrivingvehicles,wen2025ai}. Unlike traditional modular systems, VLMs benefit from their data-driven nature and end-to-end design, leading to improved generalization to novel scenarios and alleviating the problem of error propagation between modules~\cite{yang2023llm4drive, shao2023lmdriveclosedloopendtoenddriving}.

Despite their potential, current VLMs face significant obstacles in real-world deployment, particularly in latency-sensitive scenarios such as autonomous driving. The typical VLM inference pipeline involves encoding input images via a vision encoder~\cite{radford2021learning}, projecting them into the LLM-compatible embedding space, and feeding both image and text tokens into a large language model (LLM) for generation~\cite{liu2023visualinstructiontuning}. While text token generation remains relatively efficient, processing high-resolution images introduces substantial computational overhead, as it involves encoding a large number of visual tokens. These tokens, when concatenated with textual inputs, significantly increase the total sequence length, resulting in considerable inference latency due to the quadratic complexity $\mathcal{O}(N^2)$ of the attention mechanism in transformer-based VLMs~\cite{vaswani2017attention}.
This challenge becomes even more pronounced in multi-view settings, which are commonly adopted in autonomous driving systems, where synchronized images from six surrounding cameras (front, front-left, front-right, rear, rear-left and rear-right) are utilized to construct a comprehensive representation of the environment~\cite{sima2025drivelmdrivinggraphvisual, ishaq2025drivelmmo1stepbystepreasoningdataset}.
Although multi-view inputs improve safety through broader perception, they dramatically increase the number of visual tokens, leading to slower inference speeds and high memory consumption.

Among various model acceleration strategies, visual token pruning~\cite{liu2025shifting,Liu2025:GlobalCom,liu2025vidcom2,liu2025mixkv,wen2025efficient} stands out as a simple yet effective solution. Unlike quantization~\cite{gholami2022survey} or distillation~\cite{zhang2019your}, token pruning reduces token count without requiring retraining~\cite{wen2025tokenpruningmultimodallarge}. 
However, existing token pruning methods are primarily tailored for single-image inputs, neglecting not only the spatial and semantic diversity inherent in multi-view autonomous driving settings but also the fact that different camera perspectives contribute unequally to driving decisions.
Moreover, widely used token pruning strategies often depend on attention weights~\cite{wen2025tokenpruningmultimodallarge}, making them incompatible with efficient attention implementations like Flash Attention~\cite{dao2022flashattention,dao2023flashattention}, and potentially leading to suboptimal token selection in multi-view driving scenarios due to insufficient consideration of semantic coverage.

\begin{figure}[!t]
    \centering
    \includegraphics[width=1.0\linewidth]{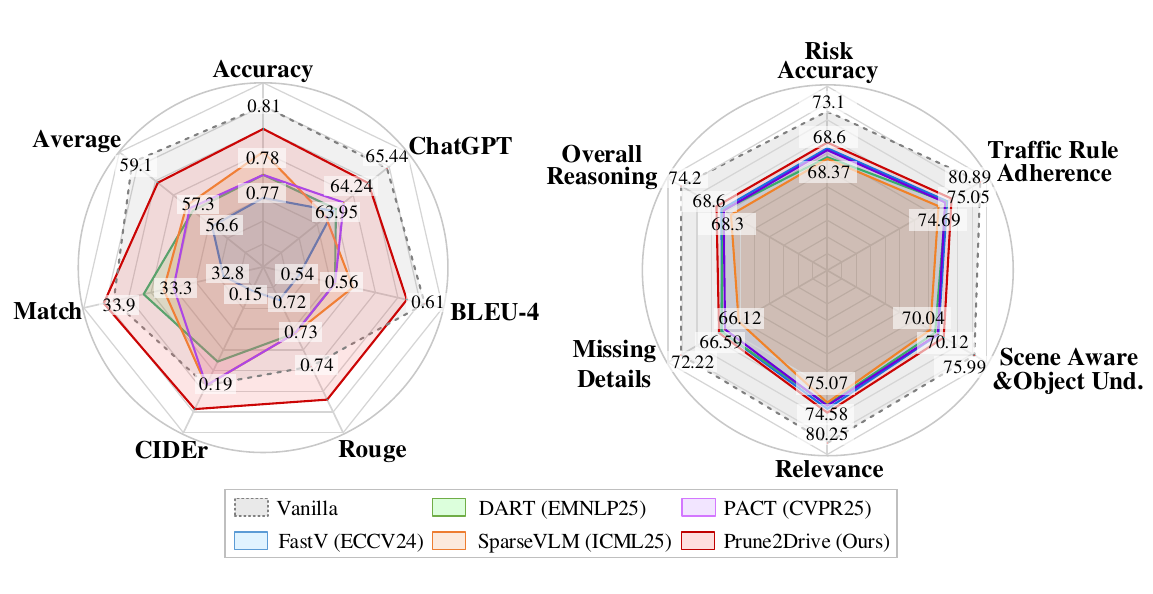}
    \caption{Prune2Drive acheives SOTA in both DriveLMM-o1 and DriveLM benchmarks. 
    }
    \label{fig:radar_fig}
    \vspace{-4mm}
\end{figure}

To bridge this gap, we introduce \textbf{\algname}. To the best of our knowledge, it is the first plug-and-play visual token pruning framework specifically designed for multi-view vision-language models in autonomous driving, addressing the unique spatial and semantic characteristics of AD scenarios.
Our method is motivated by the unique challenges of multi-view perception in autonomous driving, where indiscriminate token pruning can lead to the loss of critical spatial and semantic cues distributed across views. To this end, we introduce two key innovations in \algname.
First, we propose \textbf{T-FPS} (\textbf{T}oken-wise \textbf{F}arthest \textbf{P}oint \textbf{S}ampling), a diversity-aware token selection mechanism inspired by farthest point sampling (FPS), which selects tokens not merely based on individual salience but by maximizing semantic and spatial coverage across views. This design encourages the retention of informative but potentially low-attention tokens, ensuring holistic scene understanding even under aggressive pruning.
Second, we devise a view-adaptive pruning controller that learns to assign distinct pruning ratios to each camera view, guided by their relative contributions to downstream driving tasks. As shown in Figure~\ref{fig:teaser}, rather than relying on hand-crafted heuristics or uniform pruning, this optimization-based scheme enables our framework to allocate compute resources dynamically, preserving crucial context from front-facing or side cameras when necessary, and pruning redundant tokens elsewhere.
We validate our approach on two challenging multi-view driving benchmarks, DriveLM~\cite{huang2024drivemmallinonelargemultimodal} and DriveLMM-o1~\cite{ishaq2025drivelmmo1stepbystepreasoningdataset}, as shown in Figure~\ref{fig:radar_fig}. Experiments demonstrate that \algname consistently accelerates inference while maintaining, and in some cases even improving, task performance, highlighting its effectiveness as a plug-and-play solution for scaling vision-language reasoning in real-world AD systems.

Our contributions are summarized as follows:
\begin{itemize}
    \item We present T-FPS, a lightweight token pruning method inspired by Farthest Point Sampling that preserves semantic and spatial diversity by selecting the most representative visual tokens across multi-view inputs.
    \item We develop a view-adaptive pruning ratio optimization framework that automatically assigns distinct pruning ratios to each camera view, enabling fine-grained control over the trade-off between perception completeness and computational efficiency.
    \item We conduct comprehensive experiments on two real-world autonomous driving benchmarks, DriveLM and DriveLMM-o1. With 10\% retained tokens, our method achieves 3\% and 6\% performance drop with only 13.4\% and 20.3\% FLOPs consumption on DriveLM and DriveLMM-o1 benchmarks.
    
\end{itemize}

%% file: contents/2_related_works.tex
\section{Related Work}\label{sec:related_work}

\noindent \textbf{Vision-Language Models for Autonomous Driving.}  
Vision-language models (VLMs) have achieved remarkable success across a wide range of domains~\cite{bordes2024introduction, kim2024openvlaopensourcevisionlanguageactionmodel, yang2023llm4drive,wen2025devil,chen2024mj}. In autonomous driving (AD), VLMs can generate both high-level language instructions and low-level control commands, serving as a powerful interface for visual scene understanding and decision-making~\cite{yang2023llm4drive}.  
Recent works have explored integrating VLMs into AD systems. For example,~\cite{shao2023lmdriveclosedloopendtoenddriving} proposes an end-to-end closed-loop framework guided by natural language, while~\cite{sima2025drivelmdrivinggraphvisual} constructs a graph-based visual question answering (VQA) dataset to support multi-stage reasoning. In addition,~\cite{wang2025omnidriveholisticvisionlanguagedataset} introduces a holistic benchmark with Q-Former modules for better alignment with traditional AD tasks, and~\cite{huang2024drivemmallinonelargemultimodal} presents an all-in-one multi-modal model trained via curriculum learning to tackle diverse driving tasks.  
Despite these promising developments, VLM-based AD systems still suffer from high computational costs. The heavy visual token load, combined with multi-view inputs and large model sizes, leads to significant inference latency and GPU memory usage, posing challenges for real-time deployment and raising safety concerns in time-critical driving scenarios.

\vspace{0.3em} \noindent \textbf{Token Pruning for Vision-Language Models.}
Visual tokens are generally more redundant and less informative than text tokens~\cite{liang2022not}, making them a prime target for efficiency optimization in vision-language models. To this end, various approaches have been proposed to reduce the visual token count and accelerate inference. Some methods introduce architectural modifications by adding new modules~\cite{rao2021dynamicvitefficientvisiontransformers, li2024tokenpackerefficientvisualprojector, kim2022learnedtokenpruningtransformers}, but these typically require retraining and add complexity to the model.
In contrast, training-free token pruning~\cite{yang2025efficientvla} removes visual tokens directly during inference. This strategy is both lightweight and explainable, as it avoids additional training while maintaining interpretability.~\cite {chen2024imageworth12tokens} selects a fixed number of tokens at the second layer during inference via attention scores.~\cite{zhang2025sparsevlmvisualtokensparsification} introduces a text-guided, training-free pruning method based on cross-modal attention scores, while~\cite{liu2024multistagevisiontokendropping} proposes a multi-stage pruning framework that dynamically adjusts token selection throughout inference.
However, most existing methods rely on attention scores from a specific layer, introducing positional bias~\cite{wen2025stop} and overlooking semantically important yet low-attention tokens. Moreover, current token pruning techniques focus mainly on single-image inputs, leaving multi-view scenarios that are common in robotics and autonomous driving largely unexplored, despite their unique spatial diversity and inter-view redundancy.

%% file: contents/3_method.tex
\section{Methodology}\label{sec:method}

Figure~\ref{fig:overview} presents the overall framework of Prune2Drive through two key components to synergistically enhance efficiency while achieving optimal global resource allocation with local feature preservation: \textbf{(i) Diversity-aware token pruning}, which maintains robust scene understanding by preserving tokens that spatially cover non-redundant visual features; and \textbf{(ii) View-adaptive pruning ratio optimization}, which dynamically optimizes per-view pruning ratios by semantic importance to maximize global perception completeness, highlighting the difference in different views.

\input{contents/algorithm1}

\input{figures/overview}
\subsection{Token-wise Farthest Point Sampling}
\label{sec:method_tfps}
Inspired by the Farthest Point Sampling (FPS) algorithm, we propose T-FPS, a diversity-aware token pruning method to select non-redundant visual tokens that contain rich semantic and spatial information. FPS is a widely used sampling algorithm in point cloud processing that iteratively selects the farthest point from the already chosen set. FPS first starts by adding a randomly selected initial point to the chosen set. In each of the following iterations, FPS computes Euclidean distances between all points from the chosen set and the rest of the points, then greedily adds the most distant point from the rest of the points to the chosen set. With a designated down-sampling rate, FPS iteratively adds the farthest point to the chosen set, ensuring a trade-off between spatial distribution and geometric feature preservation.

Inspired by the coverage-maximizing principle of FPS, our method adapts this sampling philosophy to the visual token pruning of VLMs. While conventional FPS relies on geometric distances in Euclidean space, we instead employ cosine distance in the token embedding space as our primary distance metric to iteratively select a semantically-diverse and non-redundant subset of visual tokens. Such an approach effectively preserves critical semantic information by \textit{maximizing the diversity and capturing the semantic relationships between visual tokens}.

The algorithm flow of our T-FPS method is detailed in Algorithm~\ref{alg:token_pruning}. In the first phase, the algorithm initializes an empty array for the selected token subset and a distance array of length equal to the target number of retained tokens filled with infinity values, then the first token is randomly selected and appended to the selected subset. In the second phase, the algorithm performs the following steps iteratively until the selected subset reaches the target token number:

\begin{itemize}[leftmargin=10pt, topsep=0pt, itemsep=1pt, partopsep=1pt, parsep=1pt]
    \item \textbf{Pairwise Distance Computation}: Calculates the cosine distance between each unselected token and each token in the selected subset
    
    \item \textbf{Minimum Distance Preservation}: Updates each token's record in the selected subset with minimum pairwise distance to each unselected token

    \item \textbf{Farthest Token Selection}: Adds the farthest non-selected token to the selected subset
\end{itemize}

This iterative process ensures each newly added visual token maintains the largest possible distance in token-wise embedding space to the selected subset, thereby maximizing the overall semantic and spatial coverage of visual information. Our proposed T-FPS is a lightweight and efficient module which requires only a single pairwise distance computation, accounting for merely \textbf{0.02s} for N=729 tokens per image and less than 0.1\% of total FLOPS.

\subsection{View-adaptive Pruning Ratio Optimization}
\label{sec:method_ratio_opt}

In autonomous driving systems with surround-view cameras, a critical challenge is to adaptively assign distinct pruning ratios to different views. A naive, hand-crafted strategy might assign lower pruning ratios (i.e., retain more tokens) to empirically important views like the front camera. While effective, such a manual approach lacks optimality. To address this, we formulate the task of finding the best per-view ratios as a principled optimization problem.

Our goal is to automatically find an optimal vector of token retention ratios, $\boldsymbol{\alpha} = \{\alpha_1, \dots, \alpha_M\}$, for the $M$ camera views that jointly maximizes a custom objective function. This objective function is evaluated on a validation set $\mathcal{D}_{\text{val}}$ using a pre-trained and frozen VLM, denoted by $\mathcal{W}$. The optimization problem is defined as:

\begin{equation} \label{eq:search_optimized}
    \boldsymbol{\alpha}^* = \mathop{\arg\max}_{\boldsymbol{\alpha} \in \mathcal{A}} \mathcal{M}(\boldsymbol{\alpha}; \mathcal{D}_{\text{val}}, \mathcal{W})
\end{equation}
where $\mathcal{A}$ is the search space defining the permissible retention ratios for each view and potentially constrained by a pre-defined total token budget.

The objective function $\mathcal{M}$ is designed as a reward-penalty framework to explicitly balance task performance and computational efficiency:
\begin{equation}
    \mathcal{M}(\boldsymbol{\alpha}) = R(\boldsymbol{\alpha}) - \lambda P(\boldsymbol{\alpha})
\end{equation}
Where:
\begin{itemize}
    \item \textbf{Reward Term ($R$):} $R(\boldsymbol{\alpha}) = f_{\text{sim}}(y, \hat{y})$ measures the task-specific performance. It computes the language similarity between the model's output $\hat{y}$ (generated using ratios $\boldsymbol{\alpha}$) and the ground truth $y$.
    \item \textbf{Penalty Term ($P$):} $P(\boldsymbol{\alpha}) = \sum_{i=1}^{M} \alpha_i$ represents the total proportion of tokens retained across all views. This term penalizes solutions that keep too many tokens, thus encouraging sparsity and efficiency.
    \item \textbf{Trade-off Hyperparameter ($\lambda$):} $\lambda$ controls the balance between maximizing performance (reward $R$) and minimizing token usage (penalty $P$).
\end{itemize}

To solve the optimization problem in \Cref{eq:search_optimized}, we employ a parameter optimization algorithm. The search is conducted on a small, representative subset randomly sampled from the full training dataset, which serves as our validation set $\mathcal{D}_{\text{val}}$. This approach maintains sufficient data diversity for robust optimization while avoiding potential data leakage and drastically reducing the computational overhead. Notably, this entire optimization process converges efficiently within just 3 H100 GPU hours, demonstrating the practicality of our method.

\subsection{Theoretical Analysis}
\label{sec:theory}

In this section, we provide a theoretical analysis that justifies why this synergistic approach effectively minimizes information loss under a constrained budget. We begin by modeling the impact of token pruning on the model's output. We posit that the overall prediction error is a weighted aggregation of per-view information degradation.

\begin{assumption}[View-Weighted Lipschitz Continuity\label{assump:view_lipschitz}]
Let \(f\) be the end-to-end driving model, with \(y = f(\mathcal{V})\) being the output from the original multi-view token sets \(\mathcal{V} = \{V_i\}_{i=1}^M\), and \(\hat{y} = f(\mathcal{S})\) being the output from the pruned sets \(\mathcal{S} = \{S_i\}_{i=1}^M\). We assume there exist \emph{intrinsic view-importance weights} \(w_i > 0\) with \(\sum w_i = 1\), and a model-specific constant \(C_f\), such that the prediction error is bounded:
\begin{equation}
\label{equ:weighted_error}
    \|y - \hat{y}\| \le C_f \sum_{i=1}^{M} w_i \cdot d_H(V_i, S_i)
\end{equation}
where \(d_H(V_i, S_i)\) is the Hausdorff distance between the original and pruned token sets for view \(i\), defined as:
\begin{equation}
\label{equ:hausdorff_def}
    d_H(V_i, S_i) \coloneqq \max\Big\{\sup_{v \in V_i} \inf_{s \in S_i} d(v, s), \; \sup_{s \in S_i} \inf_{v \in V_i} d(s, v)\Big\}
\end{equation}
and \(d(\cdot, \cdot)\) is the cosine distance in the embedding space.
\end{assumption}
\begin{theorem}[Optimality of View-Adaptive Diversity Pruning]
\label{thm:optimality}
Under Assumption \ref{assump:view_lipschitz}, our methd, achieves a provably tighter error bound than a baseline using uniform ratios and random sampling for the same total budget \(K_{total}\):
\[
    \sum_{i=1}^{M} w_i \cdot d_H(V_i, S_{i, \text{Prune2Drive}}) \le \sum_{i=1}^{M} w_i \cdot d_H(V_i, S_{i, \text{baseline}})
\]
\end{theorem}

This result stems from our method's two-level optimization. \textbf{1)} Our T-FPS algorithm (\S\ref{sec:method_tfps}) acts as a greedy k-center solver, ensuring that the retained tokens \(S_i\) maximally cover the semantic space \(V_i\), thus minimizing the per-view information loss \(d_H(V_i, S_i)\). \textbf{2)} Our view-adaptive optimization (\S\ref{sec:method_ratio_opt}) strategically allocates larger budgets to views with higher intrinsic importance (\(w_i\)), directly minimizing the dominant terms in the total weighted error sum from \Cref{equ:weighted_error}. The combination of a superior inner-view sampling and an intelligent inter-view budget allocation jointly guarantees a lower overall error bound.


\label{sec:mmp}

%% file: contents/algorithm1.tex
\begin{algorithm}[!t]
\footnotesize
\caption{Our proposed T-FPS pruning method} 
\label{alg:token_pruning}
\begin{algorithmic}[1]
    \REQUIRE Visual token sequence $\mathbf{V} \in \mathbb{R}^{N \times d}$, target token number to be retained $\mathcal{K}$, distance measure $\mathit{dist}$
    \ENSURE Selected token indices $\mathcal{S} \subseteq \{1,\dots,K\}$ containing the most diverse and representative tokens
    \STATE \textbf{{Phase 1: Initialization}}
    \STATE Initialize distance array $\mathbf{D} \gets [+\infty]^{N}$ 
    \STATE Initialize selected token list $\mathcal{S} \gets \emptyset$ 
    \STATE $p_1 \gets \text{RandomSelect}(\{1,\dots,N\})$ 
    \STATE $\mathcal{S} \gets \mathcal{S} \cup \{p_1\}$
    
    \STATE \textbf{{Phase 2: Iteratively sampling the farthest token}}
    \FOR{$i=2$ \TO $K$} 
        \FORALL{$j \in \{1,\dots,N\} \setminus \mathcal{S}$}
            \STATE Pairwise Distance Computation: \\
            \STATE $d \gets \mathit{dist}(\mathbf{V}[j], \mathbf{V}[\mathcal{S}[i-1]])$ 
            \STATE Minimum Distance Update: $\mathbf{D}[j] \gets \min(\mathbf{D}[j], d_j)$ 
        \ENDFOR
        \STATE Farthest token selection: $p_i \gets \underset{j}{\argmax}\ \mathbf{D}[j]$ 
        \STATE Append the chosen token to the selected indices: \\
        \STATE $\mathcal{S} \gets \mathcal{S} \cup \{p_i\}$
    \ENDFOR
    
    \RETURN $\mathcal{S}$
\end{algorithmic}
\label{sec:fps}
\end{algorithm}

%% file: figures/overview.tex
\begin{figure*}[!ht]
    \centering
    \includegraphics[width=\linewidth,]{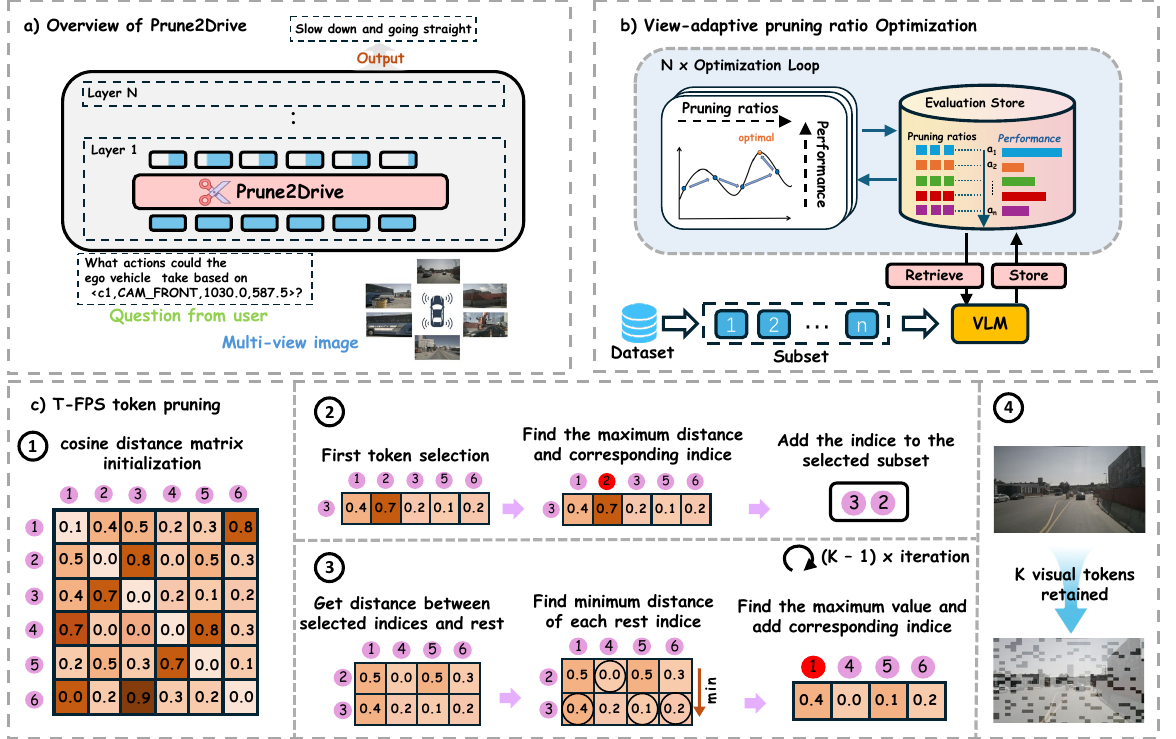}
    \vspace{-2mm}
    \caption{\textbf{Detailed architecture of Prune2Drive.} (a) VLM workflow in Prune2Drive, (b) View-adaptive pruning ratio optimization, where view-specific token pruning ratios are automatically determined, and (c) Diversity-aware T-FPS token pruning strategy, which preserves visual tokens that contain rich semantic and spatial information across multi-view inputs.}
    \vspace{-5mm}
    \label{fig:overview}
\end{figure*}

%% file: contents/4_experiments.tex
\section{Experiments}\label{sec:exp}
\input{contents/table/benchmark_Drivemm}

\input{contents/table/benchmark_drivelmm}

In this section, we evaluate our method's performance in two large autonomous driving visual-language benchmarks, which include multiple autonomous driving tasks.

\subsection{Experimental Settings}

\noindent \textbf{Implementation Details.}
We employ two specialist vision-language models: DriveMM~\cite{huang2024drivemmallinonelargemultimodal} and DriveLMM-o1~\cite{ishaq2025drivelmmo1stepbystepreasoningdataset} in two large autonomous driving benchmarks to evaluate our method's performance under various autonomous driving tasks. Specifically, DriveMM employs LLaVA-OneVision-7B~\cite{li2024llava} and fine-tunes it across different vision understanding and autonomous driving datasets, including DriveLM, to achieve specialist performance. DriveLMM-o1 employs an InternVL2.5-8B~\cite{chen2024internvl} model fine-tuned in the DriveLMM-o1 dataset, enhancing its step-by-step reasoning capability in driving scenarios.

\vspace{0.3em} \noindent \textbf{Benchmark Details.}
We use two open-source autonomous driving VQA benchmarks: DriveLM and DriveLMM-o1. DriveLM is annotated on the NuScenes~\cite{caesar2020nuscenesmultimodaldatasetautonomous} dataset, including QA pairs annotated by specialist human drivers with multi-view inputs. DriveLMM-o1 is also annotated on NuScenes, incorporating multi-view images into a step-by-step reasoning task to comprehensively evaluate quantitative decision-making capabilities in autonomous driving. We follow the common metrics in each benchmark for fair comparison: (1) DriveLM implements four evaluation metrics: accuracy, GPT score, language-based evaluation, and match score. (2) DriveLMM-o1 adopts five reasoning evaluation metrics: Risk Assessment Accuracy, Traffic Rule Adherence, Scene Awareness and Object Understanding, Relevance, and Missing Details. Furthermore, to underscore our method's scalability and effectiveness beyond multi-view scenarios, we also provide a comprehensive validation, report its performance on conventional VLM and video-based AD benchmarks. For more details on experiment settings, please refer to Suppl.~\ref{app:experiment_settings}.

\vspace{0.3em} \noindent \textbf{Comparison Methods.}
We select several recent representative methods of training-free token pruning as comparison baselines, including FastV~\cite{chen2024imageworth12tokens},   DART~\cite{wen2025stop},  SparseVLM~\cite{zhang2025sparsevlmvisualtokensparsification}, and  PACT~\cite{dhouib2025pactpruningclusteringbasedtoken}. For fair comparison, we evaluate and compare the performance of our method under two popular pruning ratio settings, where on average 75\%  and 90\% of total visual tokens are pruned.

\subsection{Main Results}

\input{contents/table/one_table_regular}
\noindent \textbf{DriveLM.} As shown in Table~\ref{tab:drivemm}, our method yield the highest in average score. We observe that \textbf{(i)} when 75\% of the visual tokens are pruned, our method achieves an impressive 58.3 average score, which only drops 1\% compared to the performance of the original model. \textbf{(ii)} Our method also demonstrates strong language-alignment with professional human-annotators, achieving the best score on all language metrics: ROUGE, CIDEr, and BLEU-4. \textbf{(iii)} Our method also demonstrates superior performance in terms of match score; This trend is particularly pronounced under moderate token-pruning settings, achieving an impressive score of 34.0 when retaining 25\% of the visual tokens, which even surpasses the original model's performance by 0.1. This proves our method's notable perception capability, enabling more accurate object identification and enhanced object detection accuracy in the driving scenario. \textbf{(iv)} Our method also leads in Accuracy Score in both token-pruning settings, which only drops 0.01 compared with the original model when retaining 25\% of the visual tokens.

\vspace{0.3em} \noindent \textbf{DriveLMM-o1.} As shown in Table~\ref{tab:drivelmm}, our method outperforms all other token-pruning methods under two pruning ratios: 75\% and 90\% on average overall reasoning ability, demonstrating its superior reasoning abilities in autonomous driving tasks within limited token counts.  It is worth noting that \textbf{(i)} baseline token pruning methods exhibit significant performance fluctuations under different pruning ratios, while our method consistently outperforms them, showing our method's robustness under diverse token budget settings. \textbf{(ii)} Across key metrics, our method leads especially in Risk assessment accuracy, Scene Awareness and Object Understanding. This trend even strengthens in aggressive token pruning settings, meaning that our method can effectively capture high-risk objects in a driving environment and make reasonable and correct judgments aligned with human drivers in dangerous scenarios. 

\vspace{0.3em} \noindent \textbf{Regular VLM benchmarks.} To assess the scalability of our T-FPS method, we evaluate across different standard VLM benchmarks. Results in Table~\ref{tab:regularvlm} highlight T-FPS's exceptional performance across varying token configurations. We observe that
\textbf{(i)} When retaining 128 tokens (22.2\% of the original), Prune2Drive achieves a remarkable \textbf{97.3\%} of the full-token performance. \textbf{(ii)} This trend even strengthens in aggressive token pruning, where our method subsubstantially outperforms second-best method, SparseVLM, by 7.5\% when 64 tokens are retained.

\vspace{0.3em} \noindent \textbf{Video AD benchmark.} To further demonstrate our method's robust performance on AD tasks with temporal input, we conduct extensive evaluation on the OmniDrive benchmark with the DriveMM model which takes multi-frame images as input. As demonstrated in Table~\ref{tab:omnidrive_comparison}, our method achieves 49.0 (vanilla: 50.3), outperforming FastV by 4.7 and SparseVLM by 2.2, demonstrating effective capture of spatial-temporal features.

\input{contents/table/vidAD}

\noindent \textbf{Inference Efficiency.} To demonstrate the efficiency of our method, we conduct a comparative analysis against two representative token pruning methods:  FastV and SparseVLM in terms of speedup in the prefilling stage and decoding stage, FLOPs, and KV cache on both DriveMM and DriveLMM-o1 models. All experiments are conducted on a single NVIDIA A100-80GB GPU. As shown in Table~\ref{tab:efficiency}, with only 10\% of the visual tokens retained, our method achieves the best efficiency scores. \textbf{(i)} Regarding CUDA latency, our method achieves 6.40$\times$ and 2.64$\times$ speedup in the prefilling stage with 1.09$\times$ and 1.04$\times$ speedup in the decoding stage, significantly improving real-world inference efficiency. \textbf{(ii)} Our method significantly reduces computational cost, achieving the lowest KV Cache and FLOPs (13.4\% and 20.3\%) compared to the original model, which achieves nearly a 9$\times$ reduction on DriveMM and 5$\times$ reduction on DriveLMM-o1. These results highlight our method's efficiency while maintaining high task performance.

\subsection{Ablation study}
\input{contents/table/ablation}

In this section, we evaluate key components of our method through ablation studies on DriveLM and DriveLMM-o1 benchmarks. We first test different optimization strategies, then explore the influence of distance measures in T-FPS token selection to assess how well they capture visual diversity and preserve critical information.

\vspace{0.3em} \noindent \textbf{Pruning Ratio Optimization Strategy.} We compare three popular and representative optimization techniques: TPE, Evolutionary, and GridSearch. The Tree-structured Parzen (TPE) is a sequential model-based optimization (SMBO) technique that uses Bayesian reasoning to efficiently explore optimization spaces. Evolutionary randomly initializes a population in the optimization space, and selects the top performers each generation. GridSearch conducts an exhaustive exploration through a manually specified subset of parameters defined in the optimization space. For fair comparison, we conduct pruning ratio optimization on the same setting in each benchmark. As shown in Table~\ref{tab:ablation-drivelmm} and Table~\ref{tab:ablation-drivemm}, TPE achieves the best performance in both DriveLM and DriveLMM-o1 benchmarks. It is worth noting that our method's performance with the optimized pruning ratio generated by GridSearch and Evolutionary is only slightly lower than TPE; this marginal performance gap further demonstrates our proposed pruning ratio optimization strategy's accuracy and robustness.
\input{figures/discussion}

\vspace{0.3em} \noindent \textbf{Distance Metrics in T-FPS.}
In our T-FPS token selection strategy, each iteration selects the token that is farthest from the already selected set, making the choice of distance metric in the token embedding space critical. Our method adopts cosine similarity as the primary distance metric. We compare it with two alternatives: $\ell_1$ and $\ell_2$ distances. As shown in Table~\ref{tab:ablation-drivelmm} and Table~\ref{tab:ablation-drivemm}, the choice of distance metric has minimal impact on overall performance. We also experiment with an inverse strategy—selecting the nearest token at each step. This results in significant performance degradation, confirming that closely clustered visual tokens are redundant and fail to capture visual diversity, which is essential for strong model performance.

\subsection{Qualitative Analysis}
Figure~\ref{fig:discussion}  highlights key differences in token selection strategy between baseline methods and our methods. Attentionbased token pruning method FastV retains numerous duplicate tokens in the back-view images, such that excessive retention of posterior visual tokens caused by positional bias
of the attention-mechanism fails to capture rich semantic
and spatial information across multi-view inputs. Similaritybased token pruning method DART uniformly selects visual tokens across different views, but fails to capture viewdependent crucial context. Our Prune2Drive considers both
the diversity of the input visual tokens to retain maximal visual information and the task-oriented contribution of different camera views to adaptively assign a visual token number to be preserved. In this example, our Prune2Drive successfully captures the crucial information in the front-view image and the back-right view image, including the white car on the left side of the front view, the black car in the middle of the front view, and the black car on the right side of the back-right view. However, both baselines fail to effectively
capture these critical objects.

%% file: contents/table/benchmark_drivemm.tex
\renewcommand{\multirowsetup}{\centering}
\begin{table}[!t]
    \centering
    \setlength{\tabcolsep}{1.2pt}
    \renewcommand{\arraystretch}{1.1}
    \footnotesize
    \centering
    \caption{Comparison with other token compression methods on DriveLM benchmark using DriveMM model.}
    \scalebox{0.80}{
    \begin{tabular}{c | c c c c c c | >{\centering\arraybackslash}p{1.0cm}}
        \toprule[1.5pt]
        \textbf{Model} & \textbf{Accuracy} & \textbf{Chatgpt} & \textbf{BLEU-4} & \textbf{Rouge} & \textbf{CIDEr} & \textbf{Match}  & {\textbf{Average}}\\
        \hline
        DriveMM & \multicolumn{7}{c}{\textit{Upper Bound, 729 Tokens per image} \ $\textbf{(100\%)}$}\\
        {Vanilla} & {0.81} & {65.44} & {0.61} & {0.74} & {0.19} & {33.9} & \multirow{1}*{{59.1}} \\
        \hline

        DriveMM & \multicolumn{7}{c}{\textit{Retain 180 Tokens per image} \ ${(\downarrow 75\%)}$}\\
        \hline
        
        FastV \texttt{\scriptsize{(ECCV24)}} & 0.77 & 63.95 & 0.54 & 0.72 & 0.15 & 32.8 & \multirow{1}*{56.6} \\
        
       DART \texttt{\scriptsize{(EMNLP25)}} & 0.78 & 64.02 & 0.56 & 0.73 & 0.18 & 33.6 & 57.2 \\
        \multirow{1}*{SparseVLM \texttt{\scriptsize{(ICML25)}}} & 0.79 & 63.74 & 0.57 & 0.73 & 0.19 & 33.4 & 57.4 \\
        PACT \texttt{\scriptsize{(CVPR25)}} & 0.78 & 64.24 & 0.56 & 0.73 & 0.19 & 33.3 & 57.3  \\
        \textbf{\algname (Ours)} & \textbf{0.80} & \textbf{64.92} & \textbf{0.60} & \textbf{0.75} & \textbf{0.20} & \textbf{34.0} & \textbf{58.3} \\
        \hline

        DriveMM & \multicolumn{7}{c}{\textit{Retain 72 Tokens per image} \ ${(\downarrow 90\%)}$}\\
        \hline
        
        FastV \texttt{\scriptsize{(ECCV24)}} & 0.68 & 63.52 & 0.49 & 0.71 & 0.08 & 32.3 & \multirow{1}*{54.1} \\
        
        DART \texttt{\scriptsize{(EMNLP25)}} & 0.69 & 64.20 & 0.51 & 0.71 & 0.12 & 32.0 & 54.7 \\
        \multirow{1}*{SparseVLM \texttt{\scriptsize{(ICML25)}}} & 0.75 & 63.38 & 0.52 & 0.72 & 0.15 & 32.9 & 55.9 \\
        PACT \texttt{\scriptsize{(CVPR25)}} & 0.76 & \textbf{64.76} & 0.54 & 0.73 & 0.15 & 32.9 & 56.8  \\

        \textbf{\algname (Ours)} & \textbf{0.78} & 64.52 & \textbf{0.56} & \textbf{0.74} & \textbf{0.16} & \textbf{33.4} & \textbf{57.4} \\

        \bottomrule[1.5pt]
    \end{tabular}}
    \label{tab:drivemm}
    \vspace{-4mm}
\end{table}

%% file: contents/table/benchmark_drivelmm.tex
\renewcommand{\multirowsetup}{\centering}
\begin{table*}[!ht]
    \centering
    \setlength{\tabcolsep}{3.5pt}
    \renewcommand{\arraystretch}{1.2}  
    \footnotesize
    \centering
	\caption{Comparison with other token compression methods on DriveLMM-o1 using DriveLMM-o1 model.}
    \scalebox{1.0}{
    \begin{tabular}{c | c c c c c| >{\centering\arraybackslash}p{1.5cm}}
        \toprule[1.5pt]
        \textbf{\begin{tabular}[c]{@{}c@{}}Methods \end{tabular}} &
  \textbf{\begin{tabular}[c]{@{}c@{}}Risk Assessment \\ Accuracy\end{tabular}} &
  \textbf{\begin{tabular}[c]{@{}c@{}}Traffic Rule\\  Adherence\end{tabular}} &
  \textbf{\begin{tabular}[c]{@{}c@{}}Scene Awareness \\ and Object Und.\end{tabular}} &
  \textbf{Relevance} &
  \textbf{\begin{tabular}[c]{@{}c@{}}Missing \\ Details\end{tabular}} &
  \textbf{\begin{tabular}[c]{@{}c@{}}Overall\\ Reasoning\end{tabular}} \\ \hline
        DriveLMM-o1 & \multicolumn{5}{c}{\textit{Upper Bound, 256 Tokens per image} \ $\textbf{(100\%)}$}\\
        {Vanilla} & {73.01} & {80.89} & {75.99} & {80.25} & {72.22} & \multirow{1}*{{74.2}} \\
        \hline

        DriveLMM-o1 & \multicolumn{5}{c}{\textit{Retain 64 Tokens per image} \ ${(\downarrow 75\%)}$}\\
        \hline
        FastV \texttt{\scriptsize{(ECCV24)}} & 68.60 & 75.05 & 70.12 & 75.07 & 66.59 & 68.6 \\
        DART \texttt{\scriptsize{(EMNLP25)}} & 67.53& 74.85& 69.59& 74.44& 66.62 & 68.4 \\ 
        \multirow{1}*{SparseVLM \texttt{\scriptsize{(ICML25)}}} & 67.33 & 73.55 & 68.80 & 74.11 & 64.23 & 67.3 \\
        
        PACT \texttt{\scriptsize{(CVPR25)}} & 68.37 & 74.69 & 70.04 & 74.58 & 66.12 & 68.3  \\
        \textbf{\algname (Ours)} & \textbf{69.28} & \textbf{75.84} & \textbf{70.99} & \textbf{75.64} &\textbf{66.95} & \textbf{69.3} \\ 
        \hline
        DriveLMM-o1 & \multicolumn{6}{c}{\textit{Retain 25 Tokens per image} \ ${(\downarrow 90\%)}$}\\
        \hline
        FastV \texttt{\scriptsize{(ECCV24)}} & 65.37 & 71.37 & 66.43 & 71.36 & 64.35 & 65.3 \\
        DART \texttt{\scriptsize{(EMNLP25)}} & 65.32& 74.62& 68.17& 73.13&65.84& 67.4 \\ 
        \multirow{1}*{SparseVLM \texttt{\scriptsize{(ICML25)}}} & 62.48 & 70.96&66.04& 70.28& 64.23& 64.8 \\ 

        PACT \texttt{\scriptsize{(CVPR25)}} & 67.01 & 73.07 & 68.42 & 72.72 & 65.53 & 67.0  \\
        \textbf{\algname (Ours)} & \textbf{68.34} & \textbf{74.92} & \textbf{69.86} & \textbf{74.54} & \textbf{66.34} & \textbf{68.3} \\

        \bottomrule[1.5pt]
	\end{tabular}}
    \label{tab:drivelmm}
     \vspace*{-0.4cm}
\end{table*}

%% file: contents/table/one_table_regular.tex
\begin{table*}[!t]
    \centering
    \begin{minipage}{0.48\textwidth}
    \centering
    \caption{Regular VLM benchmarks.}
    \vspace{-1mm}
    \setlength{\tabcolsep}{3pt}
    \renewcommand{\arraystretch}{0.9}
    \footnotesize
    \scalebox{0.92}{
    \begin{tabular}{c | c c c c c | >{\centering\arraybackslash}p{0.8cm}}
    \toprule[1.5pt]
    \textbf{Method} & \textbf{GQA} & \textbf{VQA}$^{\text{V2}}$ & \textbf{POPE} & \textbf{VizWiz} & \textbf{MME} & {\textbf{Avg}.}\\
    \hline
    LLaVA-1.5-7B & \multicolumn{6}{c}{\textit{Upper Bound, 576 Tokens}}\\
    {Vanilla} & {61.9} & {78.5} & {85.9} & {58.2} & {1506.5} & {100.0\%} \\
    \hline
    LLaVA-1.5-7B & \multicolumn{6}{c}{\textit{Retain 128 Tokens}}\\
    FastV & 54.0 & 71.0 & 68.2 & 56.4 & 1368.9 & 92.8\% \\
    SparseVLM & 57.3 & 75.1 & 83.1 & 56.3 & 1399.3 & 96.2\% \\
    \algname & \textbf{59.2} & \textbf{75.8} & \textbf{87.2} & \textbf{55.7} & \textbf{1405.3} & \textbf{97.3\%} \\
    \hline
    LLaVA-1.5-7B & \multicolumn{6}{c}{\textit{Retain 64 Tokens}}\\
    FastV & 46.0 & 55.9 & 51.6 & 35.5 & 973.5 & 74.3\% \\
    SparseVLM & 52.0 & 66.9 & 69.7 & 52.1 & 1190.4 & 86.9\% \\
    \algname & \textbf{57.6} & \textbf{74.2} & \textbf{85.4} & \textbf{54.7} & \textbf{1335.7} & \textbf{94.6\%} \\
    \bottomrule[1.5pt]
    \end{tabular}}
    \label{tab:regularvlm}
    \end{minipage}
    \hfill
    \begin{minipage}{0.48\textwidth}
        \centering
        \caption{Efficiency comparison on driving models.}
        \setlength{\tabcolsep}{2.0pt}
        \renewcommand{\arraystretch}{0.95}
        \footnotesize
        \scalebox{0.85}{
        \begin{tabular}{@{}lcccccc@{}}
            \toprule[1.2pt]
            \multirow{2}{*}{\textbf{Methods}} & 
            \multirow{2}{*}{\textbf{\#Tokens}} & 
            \multicolumn{2}{c}{\textbf{Speedup}} & 
            \multirow{2}{*}{\textbf{FLOPs}} & 
            \multirow{2}{*}{\textbf{KV Cache (MB)}} & 
            \multirow{2}{*}{\textbf{Avg}} \\
            & & \textbf{(Pre.)} & \textbf{(Dec.)} & & & \\
            \midrule
            Vanilla DriveMM & 4374 & 1.00$\times$ & 1.00$\times$ & 100\% & 2293 & 59.1 \\
            + FastV & 438 & 5.78$\times$ & 1.03$\times$ & 14.2\% & 230 & 55.4 \\
            + SparseVLM & 438 & 4.06$\times$ & 1.02$\times$ & 14.4\% & 230 & 55.9 \\
            + \algname & 438 & \textbf{6.40$\times$} & \textbf{1.09$\times$} & \textbf{13.4\%} & 230 & \textbf{57.4} \\
            \midrule
            Vanilla DriveLMM-o1 & 1536 & 1.00$\times$ & 1.00$\times$ & 100\% & 805 & 74.2 \\
            + FastV & 153 & 2.18$\times$ & 1.04$\times$ & 21.1\% & 78 & 65.3 \\
            + SparseVLM & 153 & 2.01$\times$ & 1.03$\times$ & 21.4\% & 78 & 64.8 \\
            + \algname & 153 & \textbf{2.64$\times$} & \textbf{1.04$\times$} & \textbf{20.3\%} & 78 & \textbf{68.3} \\
            \bottomrule[1.2pt]
        \end{tabular}}
        \label{tab:efficiency}
    \end{minipage}
    \vspace{-4mm}
\end{table*}

%% file: contents/table/vidAD.tex
\begin{table}[!ht]
    \centering
    \caption{Comparison on the Video AD benchmark OmniDrive.}
    \vspace{-1mm}
    \setlength{\tabcolsep}{5pt} 
    \renewcommand{\arraystretch}{1.0}
    \footnotesize
    \scalebox{0.95}{
    \begin{tabular}{l | ccc | c}
        \toprule[1.5pt]
        \textbf{Method} & \textbf{BLEU} & \textbf{CIDEr} & \textbf{ROUGE} & {\textbf{Average}}\\
        \hline
        \multicolumn{5}{c}{\textit{Baseline on OmniDrive (100\%)}} \\
        {Vanilla} & {39.1} & {77.5} & {34.2} & {50.3} \\
        \hline
        \multicolumn{5}{c}{\textit{Retain 10\% Visual Tokens} ($\downarrow$ 90\%)} \\
        FastV           & 34.5 & 68.3 & 30.1 & 44.3 \\
        SparseVLM       & 36.4 & 72.1 & 31.9 & 46.8 \\
        \textbf{Prune2Drive (Ours)} & \textbf{38.6} & \textbf{76.4} & \textbf{33.8} & \textbf{49.0} \\
        \bottomrule[1.5pt]
	\end{tabular}}
    \vspace{-2mm}
    \label{tab:omnidrive_comparison} 
\end{table}

%% file: contents/table/ablation.tex
\begin{table*}[htbp]
    \centering
    \begin{minipage}[t]{0.495\textwidth}
	\caption{Ablation Studies on DriveLMM-o1 benchmark.}
    \resizebox{0.995\textwidth}{!}{\setlength{\tabcolsep}{1.8pt}
    \renewcommand{\arraystretch}{1.272}  
    {\begin{tabular}{>{\centering\arraybackslash}m{3cm} | >{\centering\arraybackslash}m{1.8cm} >{\centering\arraybackslash}m{1.8cm} >{\centering\arraybackslash}m{2.5cm} >{\centering\arraybackslash}m{1.5cm} >{\centering\arraybackslash}m{1.8cm} | >{\centering\arraybackslash}m{1.5cm}}
        \toprule[1.5pt]
        \textbf{\begin{tabular}[c]{@{}c@{}}Model\end{tabular}} &
  \textbf{\begin{tabular}[c]{@{}c@{}}Risk  \\ Accuracy\end{tabular}} &
  \textbf{\begin{tabular}[c]{@{}c@{}}Traffic Rule\\  Adherence\end{tabular}} &
  \textbf{\begin{tabular}[c]{@{}c@{}}Scene \\ Understanding\end{tabular}} &
  \textbf{Relevance} &
  \textbf{\begin{tabular}[c]{@{}c@{}}Missing \\ Details\end{tabular}} &
  \textbf{\begin{tabular}[c]{@{}c@{}}Overall\\ Reasoning\end{tabular}} \\
        \hline
        DriveLMM-o1 & \multicolumn{6}{c}{\textit{Upper Bound, All Tokens (\textbf{100\%})} }   \\
        {Vanilla} & {73.01} &{80.89} & {75.99} & {80.25} & {72.98} & \multirow{1}*{{74.2}} \\
        \hline
        Distance Measure & \multicolumn{6}{c}{\textit{Token Reduction} \ ${(\downarrow 90 \%)}$}   \\
        \hline
        \hspace{0.2em} $\ell_1$ Distance & 66.14& \textbf{75.18}& 69.18&73.26&\textbf{66.51}& 68.3 \\
        \hspace{0.2em} $\ell_2$ Distance & 65.84&73.99& 68.84& 73.46& 65.93& 67.7  \\
        \hspace{0.2em} min Distance & 59.07& 69.10& 63.82& 67.48& 62.81& 63.0  \\
        \hspace{0.2em} cos Distance & \textbf{68.34} & 74.92 & \textbf{69.86} & \textbf{74.54} & 66.34 & \textbf{68.3}  \\
        \hline
        HPO techniques & \multicolumn{6}{c}{\textit{Token Reduction} \ ${(\downarrow 90 \%)}$}   \\
        \hline
        \hspace{0.2em} Grid Search & 65.65& 73.15& 68.62& 73.32& 66.09& 67.3 \\
        \hspace{0.2em} Evolutionary & 65.62&74.83& 68.47& 72.36& 65.82& 67.6  \\
        \hspace{0.2em} TPE & \textbf{68.34} & \textbf{74.92} & \textbf{69.86} & \textbf{74.54} & \textbf{66.34} & \textbf{68.3}  \\
        
        \bottomrule[1.5pt]
	\end{tabular}}}
    \label{tab:ablation-drivelmm}
 
    \end{minipage}
    \hfill 
    \begin{minipage}[t]{0.495\textwidth}
    
    \caption{Ablation Studies on DriveLM benchmark.}
    \resizebox{0.995\textwidth}{!}{\setlength{\tabcolsep}{1.8pt}
    \renewcommand{\arraystretch}{1.50}
    {\begin{tabular}{>{\centering\arraybackslash}m{3cm} | >{\centering\arraybackslash}m{1.8cm} >{\centering\arraybackslash}m{1.8cm} >{\centering\arraybackslash}m{1.5cm} >{\centering\arraybackslash}m{2.5cm} >{\centering\arraybackslash}m{1.5cm} >{\centering\arraybackslash}m{1.8cm} | >{\centering\arraybackslash}m{1.2cm}}
        \toprule[1.5pt]
        \textbf{\begin{tabular}[c]{@{}c@{}}Model\end{tabular}} &
        \textbf{\begin{tabular}[c]{@{}c@{}}Accuracy\end{tabular}} &
        \textbf{\begin{tabular}[c]{@{}c@{}}Chatgpt\end{tabular}} &
        \textbf{\begin{tabular}[c]{@{}c@{}}BLEU-4\end{tabular}} &
        \textbf{\begin{tabular}[c]{@{}c@{}}Rouge\end{tabular}} &
        \textbf{CIDEr} &
        \textbf{\begin{tabular}[c]{@{}c@{}}Match\end{tabular}} &
        \textbf{\begin{tabular}[c]{@{}c@{}}Average \end{tabular}} \\
        \hline
        DriveMM & \multicolumn{7}{c}{\textit{Upper Bound, All Tokens  (\textbf{100\%})} \ }   \\
        {Vanilla} & {0.81} & {65.44} & {0.61} & {0.74} & {0.19} & {33.9} & \multirow{1}*{{59.1}} \\
        \hline

        Distance Measure & \multicolumn{7}{c}{\textit{Token Reduction} \ ${(\downarrow 90 \%)}$}   \\
        \hline
        \hspace{0.2em} $\ell_1$ Distance & 0.76 & 65.04 & 0.57 & 0.74 & \textbf{0.18} & 32.7 & 56.9  \\
        \hspace{0.2em} $\ell_2$ Distance & 0.77& \textbf{65.21} & \textbf{0.59} &0.74&0.18&33.2&57.1  \\
        \hspace{0.2em} min Distance & 0.58 & 64.17 & 0.48 & 0.70 & 0.06 & 31.6 & 52.3  \\
        \hspace{0.2em} cos Distance & \textbf{0.78} & 64.52 & 0.56 & \textbf{0.74} & 0.16 & \textbf{33.4} & \textbf{57.4}  \\
        \hline
        HPO techniques & \multicolumn{7}{c}{\textit{Token Reduction} \ ${(\downarrow 90 \%)}$}   \\
        \hline
        \hspace{0.2em} Grid Search & 0.77 & 64.13 & 0.48 & 0.73 & \textbf{0.17} & \textbf{33.6} & 57.2  \\
        \hspace{0.2em} Evolutionary & 0.76 & 64.36 & 0.56 & 0.73 & 0.16 & 33.3 & 57.1  \\
        \hspace{0.2em} TPE & \textbf{0.78} & \textbf{64.52} & \textbf{0.56} & \textbf{0.74} & 0.16 & 33.4 & \textbf{57.4}  \\
        
        \bottomrule[1.5pt]
    \end{tabular}}}
    \label{tab:ablation-drivemm}
\end{minipage}
\end{table*}

%% file: figures/discussion.tex
\begin{figure*}[!t]
    \centering
    \includegraphics[width=\linewidth]{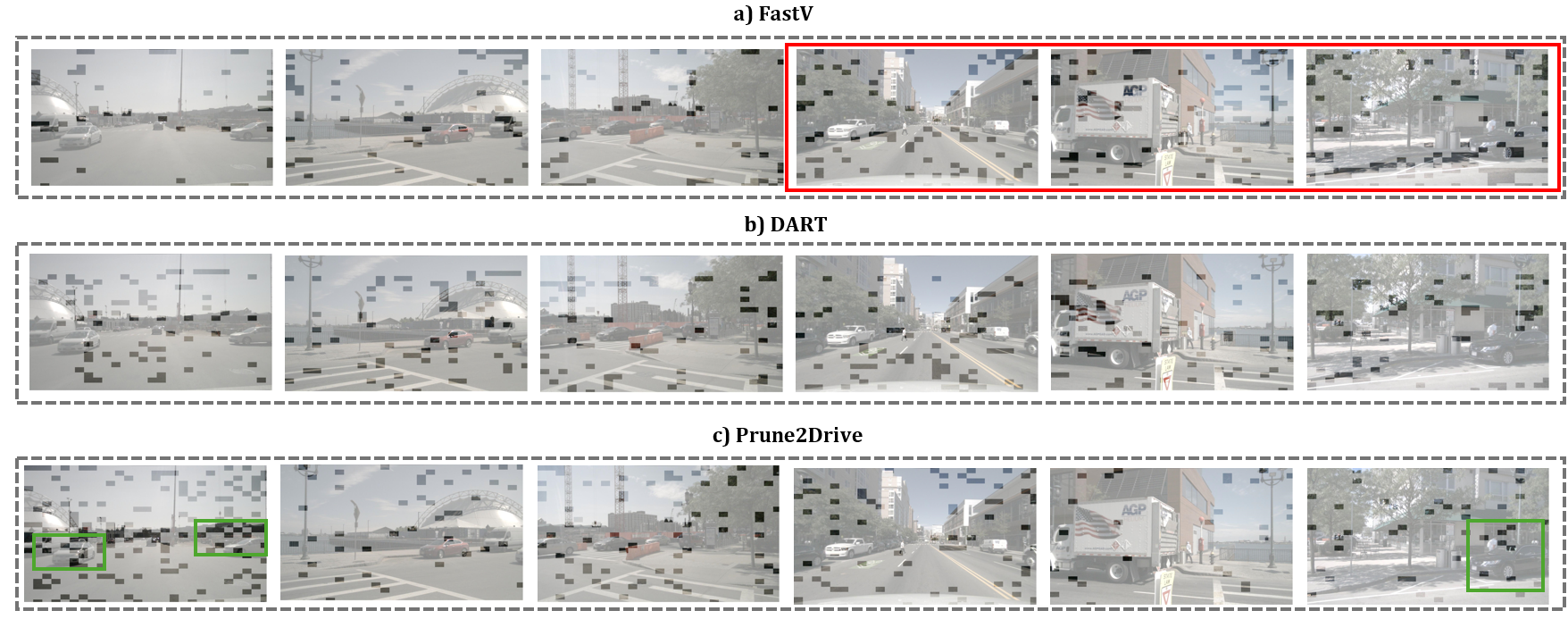}
    \vspace{-4mm}
    \caption{\textbf{Quantitative results of selected visual tokens.} We compare selected visual tokens by FastV, DART and Prune2Drive. The red box indicates the position bias of attention-based token-pruning method, where posterior tokens are retained, and the green bounding boxes highlight critical objects captured by Prune2Drive, which enables view-importance assignment and diversity-aware token selection.}
    \vspace*{-0.5cm}
    \label{fig:discussion}
\end{figure*}

%% file: contents/5_conclusion.tex
\section{Conclusion}
In this paper, we introduce \textbf{\algname}, the first plug-and-play visual token pruning framework tailored specifically for multi-view VLMs in autonomous driving scenarios. Our approach incorporates two key innovations: a diversity-aware token selection mechanism based on farthest point sampling to ensure the preservation of rich semantic information and spatial coverage among visual tokens, and a view-adaptive pruning controller that dynamically optimizes pruning ratios for each camera perspective. Extensive experiments on two large-scale autonomous driving benchmarks show that \algname achieves state-of-the-art performance across various VLM architectures. The effectiveness and scalability of our approach are further substantiated by its strong performance on both Video AD and conventional VLM benchmarks. Efficiency analysis further shows that \algname achieves a substantial reduction in computational costs while maintaining high task performance. Finally, \algname ffectively bridges the critical gap between the formidable capabilities of VLMs and practical deployment requirements, representing a significant step towards the realization of powerful, efficient, and scalable VLMs in autonomous driving systems.

%% file: contents/x_supp.tex
\clearpage
\setcounter{page}{1}
\maketitlesupplementary

\appendix
\section{Experiments Details}
\label{app:experiment_settings}

\subsection{Model architectures}
\subsubsection{DriveMM}
DriveMM adapts SigLIP as the vision encoder, which is pre-trained on WebLI with a resolution of 384×384. DriveMM uses a 2-layer MLP as the projector to project the image features into the word embedding space. For the language model, DriveMM selects Llama-3.1 8B, which utilizes a tokenizer with a vocabulary of 128,000 tokens, and the model is trained on sequences of 8,192 tokens.\\
During training, DriveMM employs a 4-stage curriculum learning approach to train the model progressively across various autonomous driving datasets, including Language-Image Alignment, Single-Image Pre-Training, Multi-Capacity Pre-Training, and Driving Fine-Tuning.

\subsubsection{DriveLMM-o1}
DriveLMM-o1 is an InternVL2.5-8B model fine-tuned on the DriveLMM-o1 dataset. InternVL2.5 consists of a Vision Transformer image encoder and an LLaMA 2-based language model, enabling strong multimodal understanding.  A key feature of InternVL2.5 is dynamic image patching, which enables variable-resolution image processing by dividing the input image into tiles, ensuring that finer details in complex scenes are better captured, which is essential for autonomous driving. \\
During training, DriveLMM-o1 freezes both the vision encoder and most of the language model’s layers, ensuring the model retains its general multimodal knowledge while adapting to domain-specific reasoning. By leveraging LoRA and dynamic image patching, this fine-tuned InternVL2.5 model effectively integrates spatial and textual reasoning, refining its ability to process diverse real-world driving situations while maintaining computational efficiency.

\subsection{Evaluation benchmarks}
To comprehensively evaluate quantitative performance in autonomous driving scenarios, we adopt the evaluation metrics proposed by both evaluation benchmarks.
\subsubsection{DriveLM}
\begin{itemize}
    \item \textbf{Accuracy}: calculates the ratio of correctly predicted samples to the total number of samples.
    \item \textbf{BLEU}: measures the similarity between a generated text and one or more
reference texts by comparing n-grams in the generated text to those in the reference texts, with higher precision indicating a better match.
    \item \textbf{ROUGE\_L}: calculates scores based on the longest common sub-sequence between the model outputs and the reference answers.
    \item \textbf{CIDEr} captures matches between n-grams of different lengths and differentiates the importance of various n-grams through TF-IDF weighting.
    \item \textbf{Chatgpt} is prompted to assign a numerical score between 0 and 100, with higher scores indicative of enhanced prediction accuracy.  The prompt we employ is as follows:
    \textit{Rate my answer based on the correct answer out of 100, with higher scores indicating that the answer is closer to the correct answer, and you should be accurate to single digits like 62, 78, 41, etc..}





    \item \textbf{Match} computes the average IoU of the predicted 2D bounding box and the ground truth 2D bounding box in the detection task.
    \item \textbf{Average Score} computes the weighted sum of Chatgpt Score, Language Score, Match Score, and Accuracy using weights of 0.4, 0.2, 0.2, and 0.2, respectively.

\end{itemize}
\subsubsection{DriveLMM-o1}
\begin{itemize}
    \item \textbf{Risk Assessment} examines the model's capacity to prioritize high-risk objects or scenarios, ensuring critical situations receive appropriate urgency in decision-making processes. Accuracy measures the precision of environmental perception by quantifying the correct identification and classification of key objects in the environment.

    \item \textbf{Traffic Rule Adherence} assesses compliance with standard traffic regulations and domain-specific best practices, mirroring real-world operational reliability.

    \item \textbf{Scene Awareness and Object Understanding} jointly evaluate how well the response interprets and reflects critical objects in current scenarios, including their positions, predictions, and actions.

    \item \textbf{Relevance} evaluates alignment between model outputs and scenario-specific requirements relative to ground truth annotations, ensuring contextually appropriate responses aligned with humans. 

    \item \textbf{Missing Details} evaluates the extent to which critical information is missing from the response by systematically analyzing perceptual gaps. 

\end{itemize}
Moreover, DriveLMM-o1 employs various complementary metrics to collectively establish a holistic framework for evaluating system reliability, safety margins, and environmental adaptability across diverse driving conditions.
DriveLMM-o1 utilizes GPT-4o-mini to complete the evaluation steps. The prompt we employ is as follows:
\textit{
You are an autonomous driving reasoning evaluator. Your task is to assess the alignment, coherence, and quality of reasoning steps in text responses for safety-critical driving scenarios. You will evaluate the model-generated reasoning using the following metrics:
}

\begin{enumerate}
    \item \textbf{Faithfulness-Step (1-10)}: Measures how well the model's reasoning steps align with the ground truth.
    \begin{itemize}
        \item 9-10: All steps correctly match or closely reflect the reference.
        \item 7-8: Most steps align, with minor deviations.
        \item 5-6: Some steps align, but several are incorrect or missing.
        \item 3-4: Few steps align; most are inaccurate or missing.
        \item 1-2: The Majority of steps are incorrect.
    \end{itemize}
    
    \item \textbf{Informativeness-Step (1-10)}: Measures completeness of reasoning.
    \begin{itemize}
        \item 9-10: Captures almost all critical information.
        \item 7-8: Covers most key points, with minor omissions.
        \item 5-6: Missing significant details.
        \item 3-4: Only partial reasoning present.
        \item 1-2: Poor extraction of relevant reasoning.
    \end{itemize}
    
    \item \textbf{Risk Assessment Accuracy (1-10)}: Evaluates if the model correctly prioritizes high-risk objects or scenarios.
    \begin{itemize}
        \item 9-10: Correctly identifies and prioritizes key dangers.
        \item 7-8: Mostly accurate, with minor misprioritizations.
        \item 5-6: Some important risks are overlooked.
        \item 3-4: Significant misjudgments in risk prioritization.
        \item 1-2: Misidentifies key risks or misses them entirely.
    \end{itemize}
    
    \item \textbf{Traffic Rule Adherence (1-10)}: Evaluates whether the response follows traffic laws and driving best practices.
    \begin{itemize}
        \item 9-10: Fully compliant with legal and safe driving practices.
        \item 7-8: Minor deviations, but mostly correct.
        \item 5-6: Some inaccuracies in legal/safe driving recommendations.
        \item 3-4: Several rule violations or unsafe suggestions.
        \item 1-2: Promotes highly unsafe driving behavior.
    \end{itemize}
    
    \item \textbf{Scene Awareness \& Object Understanding (1-10)}: Measures how well the response interprets objects, their positions, and actions.
    \begin{itemize}
        \item 9-10: Clearly understands all relevant objects and their relationships.
        \item 7-8: Minor misinterpretations, but mostly correct.
        \item 5-6: Some key objects misunderstood or ignored.
        \item 3-4: Many errors in object recognition and reasoning.
        \item 1-2: Misidentifies or ignores key objects.
    \end{itemize}
    
    \item \textbf{Repetition-Token (1-10)}: Identifies unnecessary repetition in reasoning.
    \begin{itemize}
        \item 9-10: No redundancy, very concise.
        \item 7-8: Minor repetition, but still clear.
        \item 5-6: Noticeable redundancy.
        \item 3-4: Frequent repetition that disrupts reasoning.
        \item 1-2: Excessive redundancy, making reasoning unclear.
    \end{itemize}
    
    \item \textbf{Hallucination (1-10)}: Detects irrelevant or invented reasoning steps not aligned with ground truth.
    \begin{itemize}
        \item 9-10: No hallucinations, all reasoning is grounded.
        \item 7-8: One or two minor hallucinations.
        \item 5-6: Some fabricated details.
        \item 3-4: Frequent hallucinations.
        \item 1-2: Majority of reasoning is hallucinated.
    \end{itemize}
    
    \item \textbf{Semantic Coverage-Step (1-10)}: Checks if the response fully covers the critical reasoning elements.
    \begin{itemize}
        \item 9-10: Nearly complete semantic coverage.
        \item 7-8: Good coverage, some minor omissions.
        \item 5-6: Partial coverage with key gaps.
        \item 3-4: Major gaps in reasoning.
        \item 1-2: Very poor semantic coverage.
    \end{itemize}
    
    \item \textbf{Commonsense Reasoning (1-10)}: Assesses the use of intuitive driving logic in reasoning.
    \begin{itemize}
        \item 9-10: Displays strong commonsense understanding.
        \item 7-8: Mostly correct, with minor gaps.
        \item 5-6: Some commonsense errors.
        \item 3-4: Frequent commonsense mistakes.
        \item 1-2: Lacks basic driving commonsense.
    \end{itemize}
    
    \item \textbf{Missing Step (1-10)}: Evaluates if any necessary reasoning steps are missing.
    \begin{itemize}
        \item 9-10: No critical steps missing.
        \item 7-8: Minor missing steps, but answer is mostly intact.
        \item 5-6: Some important steps are missing.
        \item 3-4: Many critical reasoning gaps.
        \item 1-2: Response is highly incomplete.
    \end{itemize}
    
    \item \textbf{Relevance (1-10)}: Measures how well the response is specific to the given scenario and ground truth.
    \begin{itemize}
        \item 9-10: Highly specific and directly relevant to the driving scenario. Captures critical elements precisely, with no unnecessary generalization.
        \item 7-8: Mostly relevant, but some minor parts may be overly generic or slightly off-focus.
        \item 5-6: Somewhat relevant but lacks precision; response contains vague or general reasoning without clear scenario-based details.
        \item 3-4: Mostly generic or off-topic reasoning, with significant irrelevant content.
        \item 1-2: Largely irrelevant, missing key aspects of the scenario, and failing to align with the ground truth.
    \end{itemize}
    
    \item \textbf{Missing Details (1-10)}: Evaluates the extent to which critical information is missing from the response, impacting the reasoning quality.
    \begin{itemize}
        \item 9-10: No significant details are missing; response is comprehensive and complete.
        \item 7-8: Covers most important details, with minor omissions that do not severely impact reasoning.
        \item 5-6: Some essential details are missing, affecting the completeness of reasoning.
        \item 3-4: Many critical reasoning steps or contextual details are absent, making the response incomplete.
        \item 1-2: Response is highly lacking in necessary details, leaving major gaps in understanding.
    \end{itemize}
\end{enumerate}

For the overall score, we compute the average of all metric scores. One example of a detailed evaluation metric is listed as follows:

\begin{verbatim}
{
  "Faithfulness-Step": 6.0,
  "Informativeness-Step": 6.5,
  "Risk Assessment Accuracy": 7.0,
  "Traffic Rule Adherence": 7.5,
  "Object Understanding": 8.0,
  "Repetition-Token": 7.0,
  "Hallucination": 8.5,
  "Semantic Coverage-Step": 7.5,
  "Commonsense Reasoning": 7.0,
  "Missing Step": 8.5,
  "Relevance": 8.5,
  "Missing Details": 7.0,
  "Overall Score": 7.42
}
\end{verbatim}
\input{figures/appendix_discussion}
\section{Comparison Baselines}
We select multiple representative training-free token pruning methods in VLMs for comparison.\\
\textbf{FastV} proposes a straightforward solution by leveraging the attention map in the second layer of LLM. It prunes tokens with the lowest visual-text attention score after layer 2 to achieve training-free token pruning. \\
\textbf{SparseVLM} adopts a multi-stage token pruning strategy. It mainly investigates the role of instruction tokens in vision-language attention mechanisms. It demonstrates that not all text tokens contribute equally to visual token pruning—only those highly relevant to the visual content are critical. To address this, the method first identifies the most vision-aligned text tokens as 'raters' and leverages their attention patterns to guide visual token pruning. \\
\textbf{DART} challenges attention-based token pruning by emphasizing redundancy reduction over token importance. It first randomly chooses pivot tokens and then selects visual tokens with minimal cosine similarity to chosen sets, thereby preventing duplication of visual information. \\
\textbf{PACT} achieves efficient visual token pruning by pruning irrelevant tokens and merging visually redundant ones at an early layer of the language model. It also proposes a novel clustering algorithm, called Distance Bounded Density Peak Clustering, which efficiently clusters visual tokens while constraining the distances between elements within a cluster by a predefined threshold.
\section{Additional Visualizations}
\subsection{Comparison with baselines}
We present additional visualizations comparing the results of retained visual tokens in Figure~\ref{fig:discussion_appendix}. It can be clearly observed that our method effectively considers the view importance while retaining informative visual tokens in aggressive pruning ratios, which is crucial for downstream autonomous driving tasks in VLMs.
\subsection{Failure analysis}
\begin{figure}[h]
\centering
\vspace{-2mm}
\includegraphics[width=0.48\linewidth, height=2.3cm]{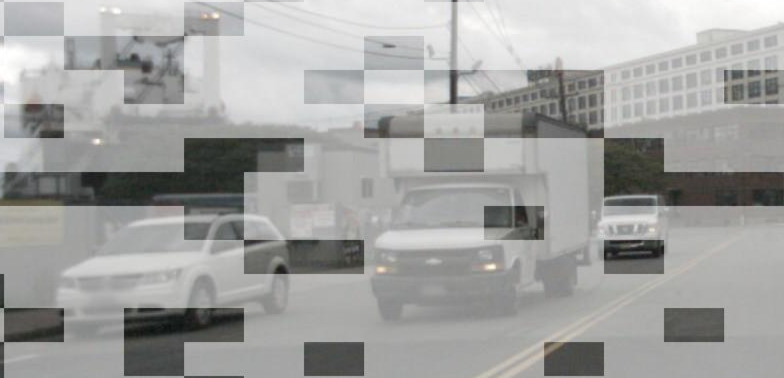}
\hfill
\includegraphics[width=0.48\linewidth, height=2.3cm]{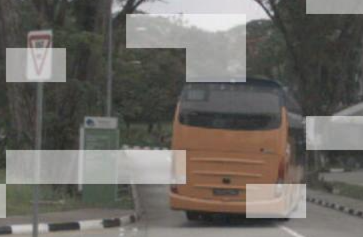}
\vspace{-2mm}
\caption{Qualitative failure cases analysis of T-FPS}
\label{fig:limitation}
\vspace{-5mm}
\end{figure}

T-FPS may under-sample safety-critical objects when they occupy large image regions with uniform color, as areas sharing similar image features lead to fewer retained tokens. Fig.~\ref{fig:limitation} illustrates two failure cases: an orange bus and three white vehicles are sparsely represented despite their safety relevance.

\section{Additional Implemention Details}
\subsection{Evaluation Details}
We introduce our evaluation details of both benchmarks in this section. For the DriveLMM-o1 benchmark, we use the official evaluation script to evaluate the full test dataset with GPT-4o-mini. For the DriveLM benchmark, following the DriveLM challenge repo, we upload our evaluation result to the official leaderboard and obtain detailed evaluation metrics after submission to the test server in Huggingface.

\subsection{Pruning ratio optimization settings}
We list our hyperparameters during the process of pruning ratio optimization below in Table~\ref{optimization_params}:
\begin{table}[H]

\caption{Hyperparameters in pruning ratio optimization}
\centering
\renewcommand{\arraystretch}{1.}
\begin{tabular}{ c c }

\hline
Hyperparameter term           & Value                            \\ 
\hline
Inital pruning ratio & {0.9}           \\
Pruning ratio upperbound    & {1}        \\
Pruning ratio lowerbound    & {0.01}       \\
Dataset size    & {500}       \\
Max iteration            & {100}     \\
Penalty scale           & {-0.05}     \\
Reward scale            & {0.5}     \\
\hline

\end{tabular}

\label{optimization_params}
\end{table}
\subsection{Hyperparameter $\lambda$ sensitivity.}
Tab.~\ref{tab:lambda} shows that similar $\lambda$ values yield comparable accuracy, demonstrating the robustness of our optimization method.

\begin{table}[h]
\centering
\small
\vspace{-2mm}
\setlength{\tabcolsep}{6pt}
\begin{tabular}{c|ccc}
\toprule
$\lambda$ & 0.04 & \textbf{0.05} & 0.06 \\
\midrule
Accuracy $\uparrow$ & 58.2 & \textbf{58.3} & 58.1 \\
\bottomrule
\end{tabular}
\vspace{-2mm}
\caption{Sensitivity of $\lambda$.}
\label{tab:lambda}
\end{table}

\subsection{Regular VLM benchmarks}
\textbf{GQA.} GQA is structured around three core components: scene graphs, questions, and images. It includes not only the images themselves but also detailed spatial features and object-level attributes. The questions are crafted to assess a model's ability to comprehend visual scenes and perform reasoning tasks based on the image content. \\
\textbf{MME.} The MME benchmark is designed to rigorously evaluate a model's perceptual and cognitive abilities through 14 subtasks. It employs carefully constructed instruction-answer pairs and concise instructions to minimize data leakage and ensure fair evaluation. This setup provides a robust measure of a model's performance across various tasks. \\
\textbf{POPE.} POPE is tailored to assess object hallucination. It presents a series of binary questions about the presence of objects in images, using accuracy, recall, precision, and F1 score as metrics. This approach offers a precise evaluation of hallucination levels under different sampling strategies. \\
\textbf{VQA V2.} VQA V2 challenges models with open-ended questions based on 265,016 images depicting a variety of real-world scenes. Each question is accompanied by 10 human-annotated answers, enabling a thorough assessment of a model's ability to accurately interpret and respond to visual queries. \\
\textbf{VizWiz.} VizWiz is a visual benchmark designed to assist visually impaired individuals. It contains real-world images captured by blind users, paired with questions they ask about the images. The dataset includes 20,523 training, 4,319 validation, and 8,000 test image-question pairs, with each question accompanied by 10 human-annotated answers. VizWiz challenges models to answer questions accurately or determine if a question is answerable, focusing on practical visual understanding and accessibility. \\

\subsection{Video AD benchmark OmniDrive}
OmniDrive is a large-scale, multi-modal dataset curated for autonomous driving VLMs. With 374,329 training and 72,184 test samples derived from nuScenes, it supports both multi-view image and video inputs. This makes it particularly suitable for tasks requiring temporal reasoning and holistic scene understanding. Its evaluation protocol utilizes rule-based language metrics for fine-grained, word-level assessment.\\
\section{Future Works}
A key limitation of this work is the lack of closed-loop evaluation. Our experiments, confined to offline AD datasets, preclude the assessment of the model's performance in dynamic, interactive environments. This fundamentally limits our understanding of its real-world applicability and long-term behavior.

%% file: figures/appendix_discussion.tex
\begin{figure*}[!t]
    \centering
    \includegraphics[width=.95\linewidth]{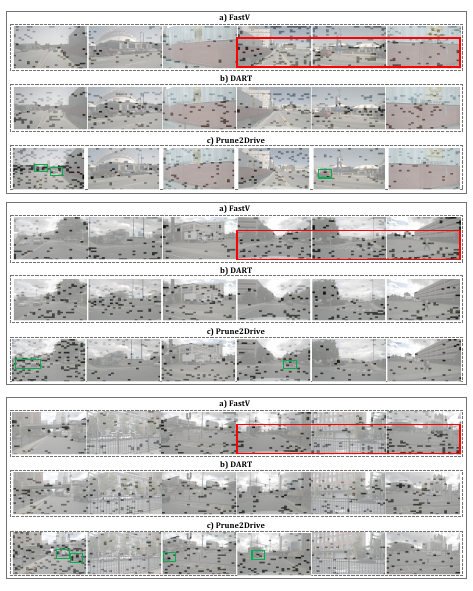}
    \vspace{-4mm}
    \caption{\textbf{Quantitative results of selected visual tokens.} We compare selected visual tokens by FastV, DART, and Prune2Drive. FastV shows position bias (red boxes), retaining mostly posterior tokens, DART neglects view importance, while our Prune2Drive (green boxes) captures critical objects through view-importance and diversity-aware selection.}
    \vspace{-4mm}
    \label{fig:discussion_appendix}
\end{figure*}

%% file: main.bib
@article{bordes2024introduction,
  title={An introduction to vision-language modeling},
  author={Bordes, Florian and Pang, Richard Yuanzhe and Ajay, Anurag and Li, Alexander C and Bardes, Adrien and Petryk, Suzanne and Ma{\~n}as, Oscar and Lin, Zhiqiu and Mahmoud, Anas and Jayaraman, Bargav and others},
  journal={arXiv preprint arXiv:2405.17247},
  year={2024}
}

@article{yang2023llm4drive,
  title={Llm4drive: A survey of large language models for autonomous driving},
  author={Yang, Zhenjie and Jia, Xiaosong and Li, Hongyang and Yan, Junchi},
  journal={arXiv preprint arXiv:2311.01043},
  year={2023}
}

@article{liu2025mixkv,
  title={Mixing Importance with Diversity: Joint Optimization for KV Cache Compression in Large Vision-Language Models},
  author={Liu, Xuyang and Gui, Xiyan and Zhang, Yuchao and Zhang, Linfeng},
  journal={arXiv preprint arXiv:2510.20707},
  year={2025}
}

@misc{kim2024openvlaopensourcevisionlanguageactionmodel,
      title={OpenVLA: An Open-Source Vision-Language-Action Model}, 
      author={Moo Jin Kim and Karl Pertsch and Siddharth Karamcheti and Ted Xiao and Ashwin Balakrishna and Suraj Nair and Rafael Rafailov and Ethan Foster and Grace Lam and Pannag Sanketi and Quan Vuong and Thomas Kollar and Benjamin Burchfiel and Russ Tedrake and Dorsa Sadigh and Sergey Levine and Percy Liang and Chelsea Finn},
      year={2024},
      eprint={2406.09246},
      archivePrefix={arXiv},
      primaryClass={cs.RO},
      url={https://arxiv.org/abs/2406.09246}, 
}

@misc{huang2024drivemmallinonelargemultimodal,
      title={DriveMM: All-in-One Large Multimodal Model for Autonomous Driving}, 
      author={Zhijian Huang and Chengjian Feng and Feng Yan and Baihui Xiao and Zequn Jie and Yujie Zhong and Xiaodan Liang and Lin Ma},
      year={2024},
      eprint={2412.07689},
      archivePrefix={arXiv},
      primaryClass={cs.CV},
      url={https://arxiv.org/abs/2412.07689}, 
}

@misc{sima2025drivelmdrivinggraphvisual,
      title={DriveLM: Driving with Graph Visual Question Answering}, 
      author={Chonghao Sima and Katrin Renz and Kashyap Chitta and Li Chen and Hanxue Zhang and Chengen Xie and Jens Beißwenger and Ping Luo and Andreas Geiger and Hongyang Li},
      year={2025},
      eprint={2312.14150},
      archivePrefix={arXiv},
      primaryClass={cs.CV},
      url={https://arxiv.org/abs/2312.14150}, 
}

@misc{shao2023lmdriveclosedloopendtoenddriving,
      title={LMDrive: Closed-Loop End-to-End Driving with Large Language Models}, 
      author={Hao Shao and Yuxuan Hu and Letian Wang and Steven L. Waslander and Yu Liu and Hongsheng Li},
      year={2023},
      eprint={2312.07488},
      archivePrefix={arXiv},
      primaryClass={cs.CV},
      url={https://arxiv.org/abs/2312.07488}, 
}

@misc{wang2025omnidriveholisticvisionlanguagedataset,
      title={OmniDrive: A Holistic Vision-Language Dataset for Autonomous Driving with Counterfactual Reasoning}, 
      author={Shihao Wang and Zhiding Yu and Xiaohui Jiang and Shiyi Lan and Min Shi and Nadine Chang and Jan Kautz and Ying Li and Jose M. Alvarez},
      year={2025},
      eprint={2405.01533},
      archivePrefix={arXiv},
      primaryClass={cs.CV},
      url={https://arxiv.org/abs/2405.01533}, 
}

@misc{chen2024imageworth12tokens,
      title={An Image is Worth 1/2 Tokens After Layer 2: Plug-and-Play Inference Acceleration for Large Vision-Language Models}, 
      author={Liang Chen and Haozhe Zhao and Tianyu Liu and Shuai Bai and Junyang Lin and Chang Zhou and Baobao Chang},
      year={2024},
      eprint={2403.06764},
      archivePrefix={arXiv},
      primaryClass={cs.CV},
      url={https://arxiv.org/abs/2403.06764}, 
}

@misc{zhang2025sparsevlmvisualtokensparsification,
      title={SparseVLM: Visual Token Sparsification for Efficient Vision-Language Model Inference}, 
      author={Yuan Zhang and Chun-Kai Fan and Junpeng Ma and Wenzhao Zheng and Tao Huang and Kuan Cheng and Denis Gudovskiy and Tomoyuki Okuno and Yohei Nakata and Kurt Keutzer and Shanghang Zhang},
      year={2025},
      eprint={2410.04417},
      archivePrefix={arXiv},
      primaryClass={cs.CV},
      url={https://arxiv.org/abs/2410.04417}, 
}

@misc{liu2024multistagevisiontokendropping,
      title={Multi-Stage Vision Token Dropping: Towards Efficient Multimodal Large Language Model}, 
      author={Ting Liu and Liangtao Shi and Richang Hong and Yue Hu and Quanjun Yin and Linfeng Zhang},
      year={2024},
      eprint={2411.10803},
      archivePrefix={arXiv},
      primaryClass={cs.CV},
      url={https://arxiv.org/abs/2411.10803}, 
}

@misc{li2024llava,
      title={LLaVA-OneVision: Easy Visual Task Transfer}, 
      author={Bo Li and Yuanhan Zhang and Dong Guo and Renrui Zhang and Feng Li and Hao Zhang and Kaichen Zhang and Peiyuan Zhang and Yanwei Li and Ziwei Liu and Chunyuan Li},
      year={2024},
      eprint={2408.03326},
      archivePrefix={arXiv},
      primaryClass={cs.CV},
      url={https://arxiv.org/abs/2408.03326}, 
}

@misc{caesar2020nuscenesmultimodaldatasetautonomous,
      title={nuScenes: A multimodal dataset for autonomous driving}, 
      author={Holger Caesar and Varun Bankiti and Alex H. Lang and Sourabh Vora and Venice Erin Liong and Qiang Xu and Anush Krishnan and Yu Pan and Giancarlo Baldan and Oscar Beijbom},
      year={2020},
      eprint={1903.11027},
      archivePrefix={arXiv},
      primaryClass={cs.LG},
      url={https://arxiv.org/abs/1903.11027}, 
}

@misc{rao2021dynamicvitefficientvisiontransformers,
      title={DynamicViT: Efficient Vision Transformers with Dynamic Token Sparsification}, 
      author={Yongming Rao and Wenliang Zhao and Benlin Liu and Jiwen Lu and Jie Zhou and Cho-Jui Hsieh},
      year={2021},
      eprint={2106.02034},
      archivePrefix={arXiv},
      primaryClass={cs.CV},
      url={https://arxiv.org/abs/2106.02034}, 
}

@misc{li2024tokenpackerefficientvisualprojector,
      title={TokenPacker: Efficient Visual Projector for Multimodal LLM}, 
      author={Wentong Li and Yuqian Yuan and Jian Liu and Dongqi Tang and Song Wang and Jie Qin and Jianke Zhu and Lei Zhang},
      year={2024},
      eprint={2407.02392},
      archivePrefix={arXiv},
      primaryClass={cs.CV},
      url={https://arxiv.org/abs/2407.02392}, 
}

@misc{kim2022learnedtokenpruningtransformers,
      title={Learned Token Pruning for Transformers}, 
      author={Sehoon Kim and Sheng Shen and David Thorsley and Amir Gholami and Woosuk Kwon and Joseph Hassoun and Kurt Keutzer},
      year={2022},
      eprint={2107.00910},
      archivePrefix={arXiv},
      primaryClass={cs.CL},
      url={https://arxiv.org/abs/2107.00910}, 
}

@article{zhou2024vision,
  title={Vision language models in autonomous driving: A survey and outlook},
  author={Zhou, Xingcheng and Liu, Mingyu and Yurtsever, Ekim and Zagar, Bare Luka and Zimmer, Walter and Cao, Hu and Knoll, Alois C},
  journal={IEEE Transactions on Intelligent Vehicles},
  year={2024},
  publisher={IEEE}
}

@article{xu2024vlm,
  title={Vlm-ad: End-to-end autonomous driving through vision-language model supervision},
  author={Xu, Yi and Hu, Yuxin and Zhang, Zaiwei and Meyer, Gregory P and Mustikovela, Siva Karthik and Srinivasa, Siddhartha and Wolff, Eric M and Huang, Xin},
  journal={arXiv preprint arXiv:2412.14446},
  year={2024}
}

@misc{jiang2024sennabridginglargevisionlanguage,
      title={Senna: Bridging Large Vision-Language Models and End-to-End Autonomous Driving}, 
      author={Bo Jiang and Shaoyu Chen and Bencheng Liao and Xingyu Zhang and Wei Yin and Qian Zhang and Chang Huang and Wenyu Liu and Xinggang Wang},
      year={2024},
      eprint={2410.22313},
      archivePrefix={arXiv},
      primaryClass={cs.CV},
      url={https://arxiv.org/abs/2410.22313}, 
}

@misc{kim2018textualexplanationsselfdrivingvehicles,
      title={Textual Explanations for Self-Driving Vehicles}, 
      author={Jinkyu Kim and Anna Rohrbach and Trevor Darrell and John Canny and Zeynep Akata},
      year={2018},
      eprint={1807.11546},
      archivePrefix={arXiv},
      primaryClass={cs.CV},
      url={https://arxiv.org/abs/1807.11546}, 
}

@misc{liu2023visualinstructiontuning,
      title={Visual Instruction Tuning}, 
      author={Haotian Liu and Chunyuan Li and Qingyang Wu and Yong Jae Lee},
      year={2023},
      eprint={2304.08485},
      archivePrefix={arXiv},
      primaryClass={cs.CV},
      url={https://arxiv.org/abs/2304.08485}, 
}

@misc{ishaq2025drivelmmo1stepbystepreasoningdataset,
      title={DriveLMM-o1: A Step-by-Step Reasoning Dataset and Large Multimodal Model for Driving Scenario Understanding}, 
      author={Ayesha Ishaq and Jean Lahoud and Ketan More and Omkar Thawakar and Ritesh Thawkar and Dinura Dissanayake and Noor Ahsan and Yuhao Li and Fahad Shahbaz Khan and Hisham Cholakkal and Ivan Laptev and Rao Muhammad Anwer and Salman Khan},
      year={2025},
      eprint={2503.10621},
      archivePrefix={arXiv},
      primaryClass={cs.CV},
      url={https://arxiv.org/abs/2503.10621}, 
}

@misc{wen2025tokenpruningmultimodallarge,
      title={Token Pruning in Multimodal Large Language Models: Are We Solving the Right Problem?}, 
      author={Zichen Wen and Yifeng Gao and Weijia Li and Conghui He and Linfeng Zhang},
      year={2025},
      eprint={2502.11501},
      archivePrefix={arXiv},
      primaryClass={cs.CL},
      url={https://arxiv.org/abs/2502.11501}, 
}

@article{vaswani2017attention,
  title={Attention is all you need},
  author={Vaswani, Ashish and Shazeer, Noam and Parmar, Niki and Uszkoreit, Jakob and Jones, Llion and Gomez, Aidan N and Kaiser, {\L}ukasz and Polosukhin, Illia},
  journal={Advances in neural information processing systems},
  volume={30},
  year={2017}
}

@inproceedings{chen2024internvl,
  title={Internvl: Scaling up vision foundation models and aligning for generic visual-linguistic tasks},
  author={Chen, Zhe and Wu, Jiannan and Wang, Wenhai and Su, Weijie and Chen, Guo and Xing, Sen and Zhong, Muyan and Zhang, Qinglong and Zhu, Xizhou and Lu, Lewei and others},
  booktitle={Proceedings of the IEEE/CVF conference on computer vision and pattern recognition},
  pages={24185--24198},
  year={2024}
}

@article{team2025kimi,
  title={Kimi-vl technical report},
  author={Team, Kimi and Du, Angang and Yin, Bohong and Xing, Bowei and Qu, Bowen and Wang, Bowen and Chen, Cheng and Zhang, Chenlin and Du, Chenzhuang and Wei, Chu and others},
  journal={arXiv preprint arXiv:2504.07491},
  year={2025}
}

@inproceedings{radford2021learning,
  title={Learning transferable visual models from natural language supervision},
  author={Radford, Alec and Kim, Jong Wook and Hallacy, Chris and Ramesh, Aditya and Goh, Gabriel and Agarwal, Sandhini and Sastry, Girish and Askell, Amanda and Mishkin, Pamela and Clark, Jack and others},
  booktitle={International conference on machine learning},
  pages={8748--8763},
  year={2021},
  organization={PmLR}
}

@article{liu2025shifting,
  title={Shifting AI Efficiency From Model-Centric to Data-Centric Compression},
  author={Liu, Xuyang and Wen, Zichen and Wang, Shaobo and Chen, Junjie and Tao, Zhishan and Wang, Yubo and Jin, Xiangqi and Zou, Chang and Wang, Yiyu and Liao, Chenfei and Zheng, Xu and Chen, Honggang and Li, Weijia and Hu, Xuming and He, Conghui and Zhang, Linfeng},
  journal={arXiv preprint arXiv:2505.19147},
  year={2025}
}

@article{Liu2025:GlobalCom,
  title={Global Compression Commander: Plug-and-Play Inference Acceleration for High-Resolution Large Vision-Language Models},
  author={Liu, Xuyang and Wang, Ziming and Han, Yuhang and Wang, Yingyao and Yuan, Jiale and Song, Jun and Zheng, Bo and Zhang, Linfeng and Huang, Siteng and Chen, Honggang},
  journal={arXiv preprint arXiv:2501.05179},
  year={2025}
}

@article{liu2025vidcom2,
  title={Video Compression Commander: Plug-and-Play Inference Acceleration for Video Large Language Models},
  author={Liu, Xuyang and Wang, Yiyu and Ma, Junpeng and Zhang, Linfeng},
  journal={arXiv preprint arXiv:2505.14454},
  year={2025}
}

@incollection{gholami2022survey,
  title={A survey of quantization methods for efficient neural network inference},
  author={Gholami, Amir and Kim, Sehoon and Dong, Zhen and Yao, Zhewei and Mahoney, Michael W and Keutzer, Kurt},
  booktitle={Low-power computer vision},
  pages={291--326},
  year={2022},
  publisher={Chapman and Hall/CRC}
}

@inproceedings{zhang2019your,
  title={Be your own teacher: Improve the performance of convolutional neural networks via self distillation},
  author={Zhang, Linfeng and Song, Jiebo and Gao, Anni and Chen, Jingwei and Bao, Chenglong and Ma, Kaisheng},
  booktitle={Proceedings of the IEEE/CVF international conference on computer vision},
  pages={3713--3722},
  year={2019}
}

@article{dao2022flashattention,
  title={Flashattention: Fast and memory-efficient exact attention with io-awareness},
  author={Dao, Tri and Fu, Dan and Ermon, Stefano and Rudra, Atri and R{\'e}, Christopher},
  journal={Advances in neural information processing systems},
  volume={35},
  pages={16344--16359},
  year={2022}
}

@article{dao2023flashattention,
  title={Flashattention-2: Faster attention with better parallelism and work partitioning},
  author={Dao, Tri},
  journal={arXiv preprint arXiv:2307.08691},
  year={2023}
}

@article{wen2025stop,
  title={Stop looking for important tokens in multimodal language models: Duplication matters more},
  author={Wen, Zichen and Gao, Yifeng and Wang, Shaobo and Zhang, Junyuan and Zhang, Qintong and Li, Weijia and He, Conghui and Zhang, Linfeng},
  journal={arXiv preprint arXiv:2502.11494},
  year={2025}
}

@article{liang2022not,
  title={Not all patches are what you need: Expediting vision transformers via token reorganizations},
  author={Liang, Youwei and Ge, Chongjian and Tong, Zhan and Song, Yibing and Wang, Jue and Xie, Pengtao},
  journal={arXiv preprint arXiv:2202.07800},
  year={2022}
}

@article{yang2025efficientvla,
  title={EfficientVLA: Training-Free Acceleration and Compression for Vision-Language-Action Models},
  author={Yang, Yantai and Wang, Yuhao and Wen, Zichen and Zhongwei, Luo and Zou, Chang and Zhang, Zhipeng and Wen, Chuan and Zhang, Linfeng},
  journal={arXiv preprint arXiv:2506.10100},
  year={2025}
}

@misc{dhouib2025pactpruningclusteringbasedtoken,
      title={PACT: Pruning and Clustering-Based Token Reduction for Faster Visual Language Models}, 
      author={Mohamed Dhouib and Davide Buscaldi and Sonia Vanier and Aymen Shabou},
      year={2025},
      eprint={2504.08966},
      archivePrefix={arXiv},
      primaryClass={cs.CV},
      url={https://arxiv.org/abs/2504.08966}, 
}

@article{team2026kimi,
  title={Kimi K2. 5: Visual Agentic Intelligence},
  author={Team, Kimi and Bai, Tongtong and Bai, Yifan and Bao, Yiping and Cai, SH and Cao, Yuan and Charles, Y and Che, HS and Chen, Cheng and Chen, Guanduo and others},
  journal={arXiv preprint arXiv:2602.02276},
  year={2026}
}

@article{wen2025ai,
  title={Ai for service: Proactive assistance with ai glasses},
  author={Wen, Zichen and Wang, Yiyu and Liao, Chenfei and Yang, Boxue and Li, Junxian and Liu, Weifeng and He, Haocong and Feng, Bolong and Liu, Xuyang and Lyu, Yuanhuiyi and others},
  journal={arXiv preprint arXiv:2510.14359},
  year={2025}
}

@article{wen2025efficient,
  title={Efficient multi-modal large language models via progressive consistency distillation},
  author={Wen, Zichen and Wang, Shaobo and Zhou, Yufa and Zhang, Junyuan and Zhang, Qintong and Gao, Yifeng and Chen, Zhaorun and Wang, Bin and Li, Weijia and He, Conghui and others},
  journal={arXiv preprint arXiv:2510.00515},
  year={2025}
}

@article{wen2026innovator,
  title={Innovator-VL: A Multimodal Large Language Model for Scientific Discovery},
  author={Wen, Zichen and Yang, Boxue and Chen, Shuang and Zhang, Yaojie and Han, Yuhang and Ke, Junlong and Wang, Cong and Fu, Yicheng and Zhao, Jiawang and Yao, Jiangchao and others},
  journal={arXiv preprint arXiv:2601.19325},
  year={2026}
}

@article{wen2025devil,
  title={The devil behind the mask: An emergent safety vulnerability of diffusion llms},
  author={Wen, Zichen and Qu, Jiashu and Liu, Dongrui and Liu, Zhiyuan and Wu, Ruixi and Yang, Yicun and Jin, Xiangqi and Xu, Haoyun and Liu, Xuyang and Li, Weijia and others},
  journal={arXiv preprint arXiv:2507.11097},
  year={2025}
}

@article{chen2024mj,
  title={Mj-bench: Is your multimodal reward model really a good judge for text-to-image generation?},
  author={Chen, Zhaorun and Du, Yichao and Wen, Zichen and Zhou, Yiyang and Cui, Chenhang and Weng, Zhenzhen and Tu, Haoqin and Wang, Chaoqi and Tong, Zhengwei and Huang, Qinglan and others},
  journal={arXiv preprint arXiv:2407.04842},
  year={2024}
}
